\documentclass[review,3p,times,authoryear]{elsarticle}

\usepackage{setspace}
\usepackage{amssymb}
\usepackage{amsmath}

\usepackage{amsthm}

\newtheorem{definition}{Definition}
\newtheorem{proposition}{Proposition}
\theoremstyle{definition}

\usepackage[ruled,linesnumbered]{algorithm2e}

\usepackage{algpseudocode}

\usepackage{multirow}
\usepackage{graphicx}

\usepackage{bm}

\usepackage[
    colorlinks=true, 
    linkcolor=blue, 
    citecolor=blue, 
    urlcolor=blue,
    pdftitle={Scalable Iterative Multi-Agent Path Finding under Turn Constraints}
    pdfauthor={Zheyu Tao, Chunlong Yu, Yicheng Hu} 
]{hyperref}

\usepackage{booktabs}

\usepackage{chngcntr}

\journal{Expert Systems With Applications}

\begin{document}

\begin{frontmatter}

\title{Scalable Iterative Multi-Agent Path Finding under Turn Constraints}

\author[Tongji University]{Zheyu Tao}
\ead{2430395@tongji.edu.cn}
\author[Tongji University]{Chunlong Yu\corref{corresponding}}
\ead{chunlong_yu@tongji.edu.cn}
\author[Tongji University]{Yicheng Hu}
\ead{2530533@tongji.edu.cn}

\cortext[corresponding]{Corresponding author.}

\affiliation[Tongji University]{
            organization={School of Mechanical Engineering, Tongji University},
            addressline={Caoan Road 4800}, 
            postcode={201804}, 
            state={Shanghai},
            country={China}
    }

\begin{abstract}

Large-scale multi-mobile-robot systems require fast and reliable path planning under realistic motion constraints. Multi-Agent Path Finding with turn actions (MAPF\textsubscript{T}) explicitly models orientation changes and turn-induced delays, but these constraints increase the computational burden of search-based methods. Priority-inheritance-based planning offers a highly scalable alternative, yet directly extending it to MAPF\textsubscript{T} is nontrivial because turn actions break the immediate node-vacation assumption underlying conventional priority inheritance. To address this challenge, this paper proposes PIBT$^{\text{*}}_{\text{T}}$, an efficient iterative solver for MAPF\textsubscript{T}. PIBT$^{\text{*}}_{\text{T}}$ advances scalable multi-agent coordination by preserving the lightweight single-timestep planning mechanism of Priority Inheritance with Backtracking (PIBT), while introducing dedicated mechanisms to handle turn-induced dependencies, local deadlocks, livelock, and inefficient repeated pushing. Experiments on standard MAPF benchmarks show that PIBT$^{\text{*}}_{\text{T}}$ achieves high success rates and strong scalability, solving instances with up to 1{,}000 agents within seconds. For realistic engineering environments, Action Dependency Graph (ADG)-based execution simulation and a flexible manufacturing system case study show that turn-aware planning better reflects robot execution and reduces actual makespan compared with classical MAPF planning. Experiments with TurtleBot4 robots further validate the practical feasibility of the proposed method.

\end{abstract}

% \begin{highlights}
% \item A turn-aware PIBT solver preserves lightweight single-timestep planning.
% \item Cycle handling, SPD, and PET address cyclic dependencies, deadlocks, and livelocks.
% \item The method solves MAPFT instances with up to 1,000 agents within seconds.
% \item ADG simulation, an FMS case study, and real-robot tests confirm execution gains.
% \end{highlights}

\begin{keyword}
Multi-mobile-robot systems, multi-agent path finding, collision avoidance, motion and path planning, turn actions
\end{keyword}

\end{frontmatter}

\section{Introduction}
\label{intro}
Multi-mobile-robot systems (MMRSs) are increasingly deployed in automated warehouses, flexible manufacturing systems, intelligent transportation, and other large-scale autonomous systems \citep{liu2023path}. Their effectiveness depends critically on the ability to coordinate large robot fleets safely and efficiently, particularly in confined environments where robot interactions are frequent and replanning must be performed online. This creates a practical demand for path-planning methods that are not only computationally scalable, but also sufficiently consistent with the motion capabilities of physical robots.

Multi-Agent Path Finding (MAPF) provides a widely used discrete framework for coordinating multiple robots \citep{stern2019multi}. Recent developments, including bounded-suboptimal conflict-based search \citep{li2021eecbs}, large-neighborhood-search methods \citep{li2022mapf,wang2025lns2+}, and LaCAM-style search \citep{okumura2023lacam,idgar2026lacam}, have enabled increasingly large MAPF instances to be solved efficiently. However, classical MAPF typically assumes that an agent can move to any neighboring vertex within one timestep, regardless of its current orientation. This abstraction ignores an important characteristic of many warehouse and logistics robots: a robot must first rotate toward its desired direction before moving forward \citep{yan2025multi}. Consequently, a path that is short under the classical MAPF metric may still incur substantial execution delay when it contains frequent direction changes.

\begin{figure*}[t]
    \centering
    \includegraphics[width=\linewidth]{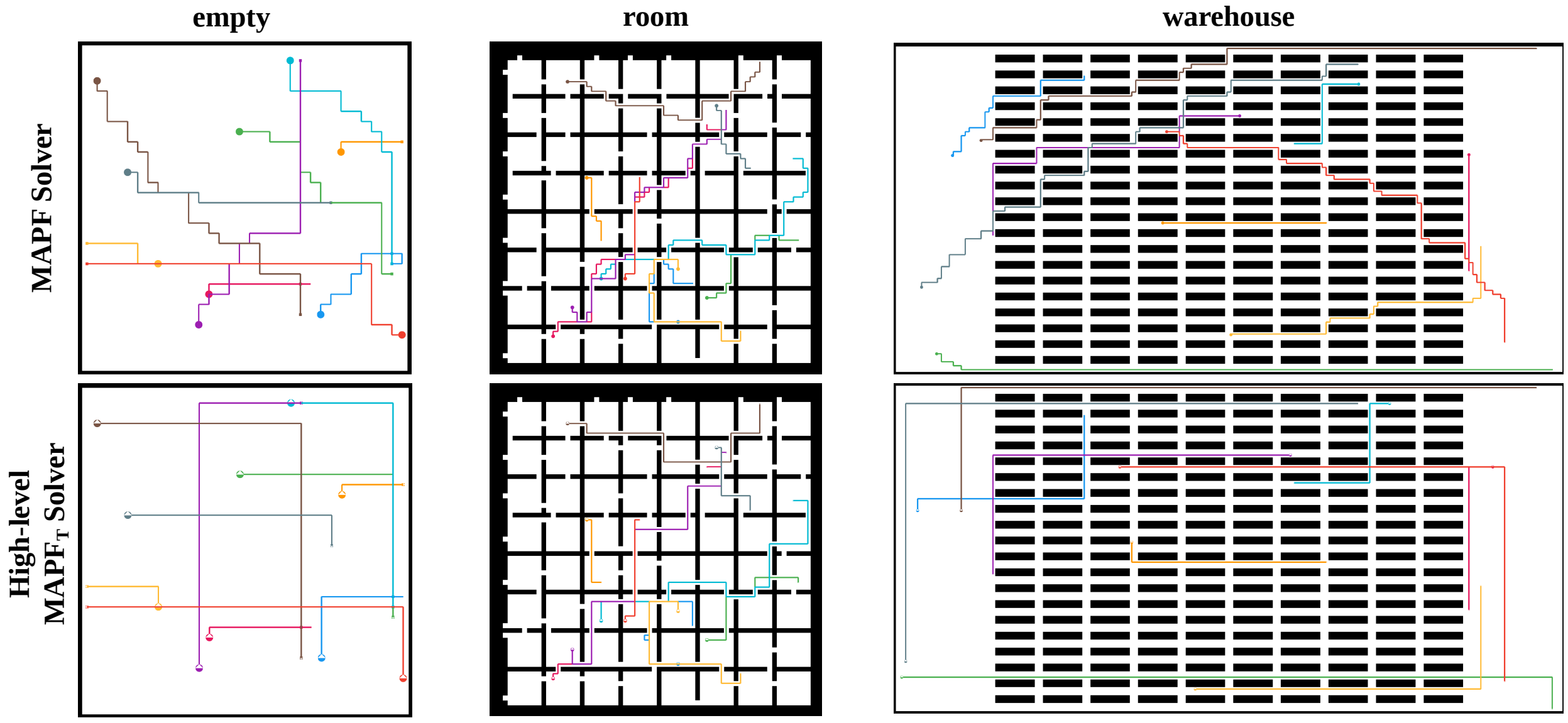}
    \caption{Paths generated by a classical MAPF solver and a high-level MAPF\textsubscript{T} solver.}
    \label{fig_path}
\end{figure*}

This discrepancy is illustrated in Fig.~\ref{fig_path}. A classical MAPF planner may favor staircase-like paths because movements in different directions are assigned identical unit costs. When executed by a mobile robot, however, each direction change introduces additional rotational, acceleration/deceleration, and synchronization delays. Explicitly modeling turn actions at the planning level can therefore produce paths that better reflect physical execution. This motivates Multi-Agent Path Finding with turn actions (MAPF\textsubscript{T}), in which orientation is included in the agent state and in-place rotations are represented explicitly. MAPF\textsubscript{T} improves motion fidelity, but the enlarged state space and additional temporal dependencies make large-scale coordination substantially more difficult.

Existing high-level MAPF\textsubscript{T} methods are predominantly search-based. Orientation-augmented ECBS formulations incorporate turning costs into bounded-suboptimal planning \citep{varambally2022mapf,honig2019persistent}, while CBS\textsubscript{T} explicitly models turn actions within conflict-based search and provides optimal solutions \citep{zhang2023efficient}. These methods demonstrate the benefit of incorporating realistic actions during high-level planning, but their computational burden increases as the number of agents and conflicts grows. This limitation is important in logistics and manufacturing applications, where hundreds or even thousands of robots may need to be coordinated within highly structured environments.

Priority Inheritance with Backtracking (PIBT) \citep{okumura2022priority} provides an attractive alternative because it generates only the next joint configuration through lightweight priority inheritance and recursive backtracking, enabling real-time coordination of very large agent populations. Extending this mechanism to MAPF\textsubscript{T}, however, is nontrivial. Conventional PIBT implicitly couples an accepted request to a neighboring vertex with the release of the agent's current vertex. Under turn constraints, these two events become temporally decoupled: an agent may successfully request a neighboring vertex but first rotate in place, leaving its current vertex occupied at the next timestep. This delayed vertex release disrupts the node-based dependency reasoning underlying priority inheritance and may create cyclic or persistent local interactions that cannot be handled effectively by conventional PIBT \citep{jiang2024scaling}.

The preliminary conference work, PIBT\textsubscript{T}, was the first, to the best of our knowledge, to extend priority-inheritance-based iterative planning to MAPF\textsubscript{T}, demonstrating that explicit turn actions can be incorporated while retaining lightweight single-timestep planning \citep{tao2025fast}. Nevertheless, the resulting local coordination remains vulnerable to several failure modes. Mutual blocking can prevent agents from making progress, repeated displacement of goal-reaching agents can delay or prevent simultaneous termination, and repeated pushing can introduce substantial detours and degrade solution quality. These limitations become increasingly important in dense interactions and restrictive map structures.

Building on this foundation, this paper develops PIBT$^{*}_{\textsubscript{T}}$, a more robust priority-inheritance-based framework for large-scale MAPF\textsubscript{T}. The proposed method preserves the computationally efficient single-timestep planning paradigm of PIBT, while strengthening its local coordination mechanisms to address deadlock, livelock, and inefficient repeated pushing. The resulting framework is evaluated not only in terms of planning performance and scalability, but also with respect to realistic robot execution through ADG-based simulation, a flexible manufacturing system case study, and physical TurtleBot4 experiments.

The main contributions of this work are summarized as follows:

\begin{enumerate}

    \item We develop PIBT$^{*}_{\textsubscript{T}}$, a robust and scalable priority-inheritance-based solver for MAPF\textsubscript{T}. The proposed framework preserves lightweight single-timestep planning while improving coordination under turn-induced temporal dependencies.

    \item We introduce two complementary coordination mechanisms to address characteristic failure modes in turn-aware priority-inheritance planning. The swap pattern detector (SPD) identifies local blocking configurations and enables cooperative yielding to resolve deadlock, while the $k$-Push Escape Trigger (PET) exploits interaction history to mitigate livelock and inefficient repeated pushing. Their individual and combined effects are systematically examined through ablation experiments.

    \item We provide a comprehensive evaluation of both algorithmic and execution-level performance. Standard MAPF benchmarks and large-scale stress tests demonstrate high success rates and scalability for up to 1{,}000 agents with limited computational overhead. ADG-based execution simulation and a flexible manufacturing system case study quantify the gap between discrete planning cost and realistic execution cost, while experiments with TurtleBot4 robots further validate the practical applicability of the proposed approach.

\end{enumerate}

The remainder of this paper is organized as follows. Section~2 reviews related work. Section~3 defines the MAPF\textsubscript{T} problem. Section~4 presents the proposed PIBT$^{*}_{\textsubscript{T}}$ framework. Section~5 reports the computational experiments and execution-oriented validation. Section~6 presents the real-robot demonstration, and Section~7 concludes the paper.

\section{Related work}
\label{sec:related_work}
Existing studies related to this work can be broadly organized along two research directions. The first incorporates realistic motion constraints, particularly orientation and turn actions, into MAPF planning and execution. The second focuses on scalable MAPF algorithms, especially priority-inheritance-based methods that enable lightweight coordination for large agent populations. We review these two directions and then position the present work at their intersection.

\subsection{MAPF with realistic motion constraints}
\label{subsec:rw_motion_constraints}

Classical multi-agent path finding (MAPF) models agents as moving between adjacent graph vertices in discrete timesteps, typically abstracting away orientation, turning time, and other motion dynamics \citep{stern2019multi}. This abstraction facilitates efficient high-level coordination, but may underestimate the execution cost of paths containing frequent direction changes. For mobile robots that move forward and rotate in place, explicitly accounting for turn actions is therefore important for obtaining plans that are temporally consistent with physical execution.

One line of research addresses realistic motion primarily at the execution level. MAPF-POST converts discrete MAPF paths into executable schedules subject to kinematic constraints \citep{honig2016multi}, while the action dependency graph (ADG) represents precedence relationships between robot actions to coordinate robust execution \citep{honig2019persistent}. Subsequent studies improve execution flexibility by relaxing unnecessary dependencies, changing passing orders during execution, or reducing redundant waiting \citep{berndt2023receding,su2024bidirectional,dunkel2024streamlining}. The SMART testbed and subsequent analyses further show that the performance of a multi-robot system depends jointly on the high-level planner and the execution framework \citep{yan2026analyzing, yan2026advancing}. These approaches demonstrate that a discrete MAPF solution can be converted into a physically executable motion schedule. However, execution-level adjustment does not ensure that the high-level routes themselves account for the temporal cost of turning and may lead to worse solution quality \citep{yan2026advancing}. This distinction motivates MAPF formulations that incorporate realistic actions directly into the planning process.

A second line of research explicitly models orientation and turning during high-level planning. Orientation-augmented ECBS formulations provide bounded-suboptimal planning with turn actions, and computational studies demonstrate the benefit of accounting for rotations in warehouse environments \citep{varambally2022mapf}. Their action abstraction, however, adopts a simplified representation of rotational motion and does not fully distinguish the execution costs associated with turns of different angular magnitudes. CBS\textsubscript{T} further develops an optimal solver for MAPF\textsubscript{T} by incorporating turn actions into conflict-based search together with symmetry reasoning and search acceleration \citep{zhang2023efficient}. Other studies consider richer motion models using different planning paradigms. TP-SIPPwRT combines token passing with safe-interval planning for kinematically constrained pickup-and-delivery tasks and provides completeness for well-formed instances \citep{ma2019lifelong}. A three-level stationary-state search called MASS models orientation, velocity, and acceleration for differential-drive robots \citep{yan2025multi}. Collectively, these studies establish the value of explicitly modeling robot motion, although their formulations, guarantees, and demonstrated scales differ substantially.

MAPF\textsubscript{T} introduces difficulties beyond the increase in state-space size. In particular, a rotation may consume time while the agent continues to occupy its current vertex. Consequently, an intended transition to a neighboring vertex does not necessarily coincide with the release of the currently occupied vertex. Such delayed vertex release introduces additional temporal dependencies into multi-agent coordination and complicates conflict resolution. In global search frameworks such as CBS\textsubscript{T}, dense interactions may further require extensive exploration of the constraint tree. A key challenge is therefore to retain explicit turn modeling while coordinating large agent populations without incurring excessive global-search overhead.

\subsection{Scalable and priority-inheritance-based MAPF}
\label{subsec:rw_scalable_mapf}

MAPF scalability has been pursued through several complementary strategies. EECBS accelerates conflict-based search while retaining a prescribed suboptimality bound \citep{li2021eecbs}. Large-neighborhood-search methods improve feasible plans by repeatedly replanning selected subsets of agents, while MAPF-LNS2 focuses on conflict repair and LNS2+RL incorporates learned coordination into the repair process \citep{li2021anytime,li2022mapf,wang2025lns2+}. For lifelong settings, RHCR reduces planning complexity through temporal decomposition, repeatedly solving MAPF over a limited planning horizon \citep{li2021lifelong}. These approaches improve scalability by restricting the search space, replanning scope, or temporal horizon while retaining a search- or replanning-based coordination structure.

Priority-inheritance-based planning provides a different approach to large-scale coordination. PIBT constructs the next joint configuration using dynamic priorities, priority inheritance, and recursive backtracking, enabling highly efficient local coordination without explicitly constructing a large global search tree \citep{okumura2022priority}. The same mechanism also serves as an efficient successor generator within higher-level search frameworks. LaCAM performs search over joint configurations while using constrained PIBT to generate successor configurations \citep{okumura2023lacam}. LaCAM* extends this framework toward eventual optimality for makespan and sum of loss, and subsequent variants further improve makespan-oriented search \citep{okumura2023improving,okumura2024engineering,idgar2026lacam}. Real-Time LaCAM maintains such search incrementally across execution steps while preserving completeness \citep{liang2025real}. These developments show that lightweight PIBT-based configuration generation can support both direct iterative coordination and more systematic global search.

For PIBT used directly as an iterative planner, however, its behavior depends strongly on the structural assumptions underlying local configuration generation. Existing extensions address different limitations of the original framework. WinPIBT enlarges the planning window \citep{okumura2019winpibt}, Causal-PIBT introduces asynchronous action dependencies \citep{okumura2021time}, and temporary-priority mechanisms mitigate deadlocks in maps containing tree-like regions \citep{fujitani2022deadlock}. Improved preference construction further enhances local action selection \citep{okumura2025lightweight}, while Anytime PIBT shows that improving individual single-step decisions does not necessarily lead to comparable improvements in complete solution paths \citep{gandotra2025anytime}. These extensions retain the efficiency of local priority inheritance while addressing specific coordination limitations.

Introducing turn actions poses a more fundamental difficulty for conventional PIBT reasoning. In the classical action model, accepting a request to move to a neighboring vertex is closely associated with vacating the current vertex. With explicit rotations, an agent may instead accept such a request, rotate first, and continue occupying its current vertex for one or more action steps. The intended movement and the actual release of the occupied vertex are therefore temporally decoupled. This mismatch complicates the node-based dependency reasoning on which priority inheritance and recursive backtracking rely \citep{jiang2024scaling}.

\subsection{Positioning of the present work}
\label{subsec:rw_positioning}
Taken together, the above studies reveal a gap between realistic turn-aware planning and scalable local coordination. Existing search-based MAPF\textsubscript{T} methods explicitly model turn actions, but their computational burden can grow substantially with the number of agents and conflicts. In contrast, PIBT provides highly efficient priority-inheritance-based coordination, yet its node-based reasoning becomes nontrivial under turn-induced delayed vertex release.

To the best of our knowledge, the preliminary conference work, PIBT\textsubscript{T}, was the first to extend priority-inheritance-based iterative planning to MAPF\textsubscript{T}, demonstrating that lightweight single-timestep planning can be retained under explicit turn constraints \citep{tao2025fast}. However, the resulting coordination mechanism remains vulnerable to local blocking, repeated displacement of goal-reaching agents, and inefficient pushing behaviors, particularly in dense interactions and restrictive map structures. Building on this foundation, the present study develops PIBT$^{*}_{\textsubscript{T}}$, a more robust turn-aware priority-inheritance framework for large-scale MAPF\textsubscript{T}. The proposed method strengthens local coordination by explicitly addressing the failure modes that remain in PIBT\textsubscript{T}, while preserving its lightweight single-timestep planning paradigm.

\section{Problem description} \label{sec:problem}

\textit{1) Input: } 
MAPF\textsubscript{T} considers a set of agents \( \bm{A} = \{ a_1, a_2, \ldots, a_k \} \) operating on an undirected grid graph \(\bm{G}=(\bm{V},\bm{E})\).  The initial configuration of each agent is defined by the tuple \(s_i[0]=(\pi_i[0],\theta_i[0])\), which consists of its starting node \(\pi_i[0]\in \bm{V}\) and initial orientation \(\theta_i[0]\in \{North, East, South, West\}\). Each agent is further assigned a specific goal node \(g_i\in \bm{V}\), and no terminal orientation is required.

\textit{2) Action model: }
At each timestep, an agent can execute one of three actions: move forward to the adjacent node in its current orientation, rotate by 90 degrees clockwise or counterclockwise while remaining in place, or wait in place without changing its orientation.

\textit{3) Conflict: }
At any timestep, a conflict occurs if two or more agents attempt either of the following: occupy the same node, which constitutes a \textbf{vertex conflict}; or traverse the same edge in opposite directions, which constitutes an \textbf{edge conflict}.

\textit{4) Solution: }
A solution to MAPF\textsubscript{T} is a set of action sequences, one for each agent, that move the agents from their initial states to their respective goal nodes while ensuring that the entire process is free of conflicts.

\textit{5) Performance metrics: }
The quality of an MAPF\textsubscript{T} solution can be evaluated using the following metrics: The \textbf{Sum of costs (SoC)} is defined as the total number of timesteps taken by all agents from their initial states until they reach and remain at their respective goal nodes.  The \textbf{makespan} is the elapsed time until all agents have reached their goal nodes, which is equivalently the timestep at which the last agent arrives at its goal.

\section{Methodology}\label{sec:Method}
This section presents the proposed PIBT$^{\text{*}}_{\text{T}}$ method for MAPF\textsubscript{T}. The method aims to preserve the lightweight single-timestep planning structure of PIBT while extending it to an orientation-aware action model. We first review the original PIBT framework and explain why its immediate node-vacation assumption becomes invalid under turn constraints. We then introduce the operational mechanism of PIBT$^{\text{*}}_{\text{T}}$, including the recursive priority inheritance procedure and the turn-aware action update function. Next, we present several dedicated mechanisms for addressing the algorithmic difficulties that arise in MAPF\textsubscript{T}, including cycle handling for circular request chains, a local greedy strategy for orientation-aware node selection, a swap pattern detector for local deadlock resolution, and a $k$-push escape trigger for mitigating livelock and inefficient repeated pushing behaviors. Finally, we analyze the computational complexity of the proposed algorithm.

\subsection{PIBT}
PIBT is built around a simple priority-based planning mechanism. At each timestep, agents are ordered according to their priorities and plan their next configurations sequentially from the highest priority to the lowest priority. Each agent selects a preferred node from its candidate set, typically consisting of its current node and neighboring nodes, and attempts to reserve that node for the next timestep. Through this sequential reservation process, PIBT incrementally constructs a conflict-free joint configuration for the next timestep.

The core mechanism of PIBT is to resolve conflicts through priority inheritance. 
When an agent \( a_i \) requests a node that is currently occupied by a lower-priority agent \( a_j \), 
\( a_i \) temporarily transfers its priority to \( a_j \) and forces \( a_j \) to move away from the requested node. 
If \( a_j \) is itself blocked by another lower-priority agent, the same process is applied recursively, 
forming a chain of priority inheritance until all blocking agents attempt to relocate.

This simple configuration-generation mechanism gives PIBT very low computational overhead, enabling it to solve large-scale MAPF instances efficiently. However, directly applying PIBT to MAPF\textsubscript{T} is nontrivial. In MAPF\textsubscript{T}, an agent may need to rotate in place before moving into a requested neighboring node. Consequently, a successful node request does not necessarily imply that the agent's current node will be vacated at the next timestep. This violates the immediate node-vacation assumption underlying the recursive conflict-resolution logic of PIBT. To address this limitation, we propose PIBT$^{\text{*}}_{\text{T}}$.

\subsection{Operational mechanism of \texorpdfstring{$\text{PIBT}^{\text{*}}_{\text{T}}$}{PIBT*T}}
\label{sec:PIBTT*}
Before presenting the detailed procedure, we highlight how PIBT$^{\text{*}}_{\text{T}}$ modifies the original PIBT mechanism. In PIBT, node reservation and next-step occupation are tightly coupled: once a node request is accepted, the corresponding agent is assumed to move into the requested node at the next timestep. PIBT$^{\text{*}}_{\text{T}}$ decouples these two operations. The recursive priority-inheritance procedure is still used to determine requested nodes, but each accepted request is converted into an executable next configuration by a turn-aware action update function. This design allows an agent to rotate, move forward, or wait according to its current orientation, while additional checks are introduced to handle circular request chains caused by turn-induced delays.

\begin{algorithm}[t]
    \caption{PIBT$^{\text{*}}_{\text{T}}$ at timestep $t$}
    \label{algorithm}
    \LinesNumbered
    \KwIn{Graph $G$, agents $A$,  current configuration  \(Q[t]=\{(\pi_1[t],\theta_1[t]),\ldots,(\pi_k[t],\theta_k[t])\}\)}
    \KwOut{new configuration \(Q[t+1]\)} 
    Update priorities $p_i$ for each agent \(a_i\in \bm{A}\)\;
    Sort $A$ by updated priorities\;
    \ForEach {\textnormal{agent} \(a_i\in \bm{A}\)}{
        %\If{\(\pi_i[t+1]=\) \textnormal{NULL}\textbf{\textup{ and }}\(\theta_i[t+1]=\)\textnormal{NULL}}{
        \If{\(Q_i[t+1]=\) \textnormal{NULL}}{
            PIBT$^{\text{*}}_{\text{T}}$\((a_i,\text{NULL})\)\;  
            }
    }
    \BlankLine

    \SetKwFunction{MyFunction}{PIBT$^{\text{*}}_{\text{T}}$}
    \SetKwProg{Fn}{function}{:}{}
    \Fn{\MyFunction{$a_i,a_j$}}{
        $P_i \leftarrow Neighbor(\pi_i[t])\cup \pi_i[t]$\;
        Sort $P_i$ by \textbf{\textup{node selection strategy}}\; \label{Alg1, node_selection}
        \ForEach{$u\in P_i$}{
            \If{$\exists a_k\in A, \pi_k[t+1]=u$ \textbf{\textup{ or }} $a_j \neq \textnormal{NULL} \bigwedge \pi_j[t]=u$}{ \label{line:start_conflict}
                continue\;
            } \label{line:end_conflict}
            \If{$a_i$ \textnormal{is involved in }cycle}{ \label{line:start_cycle}
                Determine $Q_i[t+1]$ by cycle handling approach\;
                \KwRet{\textnormal{True}}\;
            } \label{line:end_cycle}
            \If{$\exists a_k\in A,\pi_k[t]=u \bigwedge \pi_k[t+1]=\textnormal{NULL}$}{ \label{Alg1, rec_check}
                \If{\MyFunction{$a_k,a_i$}=\textnormal{False}\label{Alg1, rec}}{continue\;} 
            }
            \If{$Q_i[t+1]$ \textnormal{has been determined in cycle handling approach}}{
                \KwRet{\textnormal{True}}\;
            }
            % $(\pi_i[t+1],\theta_i[t+1])\leftarrow \textnormal{Compute\_{ }action}(\pi_i[t],\theta_i[t]),u)$\; \label{Alg1, compute_action}
            
            % \If{$\exists a_k\in A,\pi_k[t+1]=\pi_i[t+1]$}{ \label{line:start_update_wait}
            %     $\pi_i[t+1]\leftarrow \pi_i[t]$\;
            % }  \label{line:end_update_wait}
            $Q\textsubscript{next}$ $\leftarrow \textnormal{Compute\_Action}((\pi_i[t], \theta_i[t]), u)$\; \label{Alg1, compute_action}
            \uIf{$\exists a_k \in A, \pi_k[t+1] = \pi_{next}$}{ \label{line:check_conflict}
                %$\pi_i[t+1] \leftarrow \pi_i[t], \theta_i[t+1] \leftarrow \theta_i[t]$\; \label{line:update_as_wait}
                $Q_i[t+1] \leftarrow Q_i[t]$\;  \label{line:update_as_wait}
            }
            \Else{$Q_i[t+1] \leftarrow Q\textsubscript{next}$\; \label{line:commit}}
            \KwRet{\textnormal{True}}\;
        }
        $Q_i[t+1] \leftarrow Q_i[t]$\; \label{Alg1, invalid}
        \KwRet{\textnormal{False}}\;
    } 
\end{algorithm}

Algorithm~\ref{algorithm} describes one planning step of PIBT$^{\text{*}}_{\text{T}}$ at timestep \(t\). Before the iterative planning process starts, each agent is assigned a unique initial priority value in \((0,1)\), which is used as a tie-breaker. At the beginning of each timestep, priorities are updated according to the same rule as in PIBT: the priority of an agent that has not reached its goal is increased, whereas the priority of an agent already at its goal is reset to its initial value. Agents are then processed in descending order of priority. If the next configuration of agent \(a_i\) has not yet been determined, the recursive procedure PIBT$^{\text{*}}_{\text{T}}(a_i,\textnormal{NULL})$ is invoked.

The recursive procedure PIBT$^{\text{*}}_{\text{T}}(a_i,a_j)\) determines a feasible next configuration for agent \(a_i\), where \(a_j\) denotes the parent agent that has requested \(a_i\) to move, if such an agent exists. The procedure first constructs the candidate set \(P_i\), which consists of the current node of \(a_i\) and its neighboring nodes. These candidates are ordered by the node selection strategy introduced in Section~\ref{Sec. Node_Selection}. The algorithm then examines the candidates sequentially according to this preference order.

For a candidate node \(u\), the algorithm first checks whether \(u\) has already been reserved by another agent at timestep \(t+1\), or whether \(u\) is the current node of the parent agent \(a_j\). The former condition prevents two agents from reserving the same next-step node, while the latter avoids an immediate swap with the parent agent. If either condition holds, candidate \(u\) is skipped.

If \(u\) is currently occupied by another agent \(a_k\) whose next configuration has not yet been determined, priority inheritance is triggered by recursively calling PIBT$^{\text{*}}_{\text{T}}(a_k,a_i)\). In this way, \(a_k\) temporarily inherits the priority of \(a_i\) and attempts to determine its own next configuration first. If the recursive call fails, \(a_i\) continues to test the next candidate node in \(P_i\).

In addition to the standard recursive inheritance process, PIBT$^{\text{*}}_{\text{T}}$ checks whether the current node request closes a circular request chain. If such a cycle is detected, the cycle-handling mechanism is invoked to determine consistent next-step configurations for all agents involved in the cycle. The details of this mechanism are given in Section~\ref{Sec. Cycle}.

Once a candidate request is accepted, PIBT$^{\text{*}}_{\text{T}}$ computes the next configuration by calling the Compute\_action function (Section \ref{Sec. action_update}). This function converts the accepted node request into the first executable action under the MAPF\textsubscript{T} action model. The resulting configuration is denoted by \(Q_{\mathrm{next}}=(\pi_{\mathrm{next}},\theta_{\mathrm{next}})\). If \(\pi_{\mathrm{next}}\) has already been reserved by another agent at timestep \(t+1\), agent \(a_i\) waits and keeps its current configuration. Otherwise, \(Q_{\mathrm{next}}\) is assigned to \(Q_i[t+1]\).

If no candidate node in \(P_i\) yields a feasible update, agent \(a_i\) waits in place, i.e., \(Q_i[t+1]\leftarrow Q_i[t]\), and the function returns \textnormal{False}. The parent call may then continue to test other candidate nodes. Through this recursive process, PIBT$^{\text{*}}_{\text{T}}$ incrementally constructs a conflict-free joint configuration for all agents at timestep \(t+1\). This one-step planning procedure is repeated over successive timesteps until all agents reach and remain at their goal nodes.

\begin{figure}[t]
	\centering
	\includegraphics{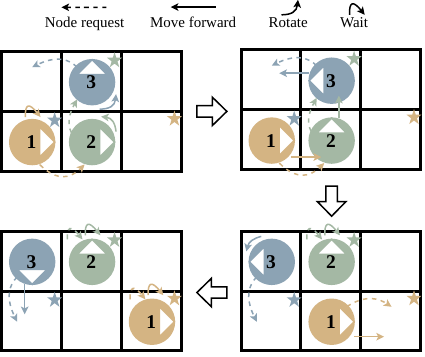}
	\caption{An MAPF\textsubscript{T} instance solved by the proposed method.}
	\label{fig_ins}
\end{figure}

\subsection{Action update function} \label{Sec. action_update}

Given the current state \((\pi_i[t],\theta_i[t])\) of agent \(a_i\) and a requested node \(u\), the Compute\_action function (line \ref{Alg1, compute_action}, Algorithm \ref{algorithm}) determines the first executable action toward \(u\) as follows:

\begin{enumerate}
\item If the requested node lies in the agent’s current facing direction, the agent moves forward.
\item If the requested direction is 90° to the left or right of the current orientation, the agent performs a single 90° rotation toward that direction.
\item If the requested direction is opposite (180° difference), the agent performs a 90° counterclockwise rotation.
\item If the requested node is the agent’s current node, the agent remains in place and executes no action.
\end{enumerate}

This function is the key component that decouples node requests from actual motion execution. A requested neighboring node represents the intended movement direction, whereas the executable next configuration may be a forward movement, an in-place rotation, or a wait action. The following example illustrates how node requests and action updates interact in PIBT$^{\text{*}}_{\text{T}}$.

\subsection{Illustrative example}
An illustrative example of the basic planning process of PIBT$^{\text{*}}_{\text{T}}$ is shown in Figure~\ref{fig_ins}. The example involves three agents, \(a_1\), \(a_2\), and \(a_3\), on a \(2\times 3\) grid with assumpted priority order \(a_1>a_2>a_3\). In the figure, the goal nodes are represented by stars of the corresponding colors. Dashed arrows indicate successful node requests, while solid arrows represent the actual actions executed in the current timestep.

At the initial timestep, agent \(a_1\), as the highest-priority agent, first requests the node currently occupied by \(a_2\). This triggers the recursive procedure PIBT$^{\text{*}}_{\text{T}}(a_2,a_1)\). Agent \(a_2\) then requests the node occupied by \(a_3\), which in turn triggers PIBT$^{\text{*}}_{\text{T}}(a_3,a_2)\). Agent \(a_3\) successfully requests the node to its left, which is currently unoccupied. The action update function therefore assigns \(a_3\) a \(90^\circ\) counterclockwise rotation as its first action toward the requested node. This updated information is then returned to \(a_2\). Following the same logic, \(a_2\) also performs a \(90^\circ\) counterclockwise rotation. Although \(a_1\) successfully requests the node to its right and is already oriented toward that direction, the requested node remains blocked because \(a_2\) is still rotating in place. As a result, \(a_1\) waits during the current timestep.

At the next timestep, the node request process remains unchanged because the positions of all agents are the same as in the previous timestep. However, after the rotations completed in the previous step, \(a_3\) can now move one step to the left. Once the updated action information of \(a_3\) is returned, \(a_2\) can move upward, and \(a_1\) can move to the right. Hence, all three agents make progress in this timestep.

During the third timestep, the algorithm again plans actions according to the updated priority order. Agent \(a_1\) first requests the goal node on its right. Since this node is unoccupied and \(a_1\) is already aligned with it, \(a_1\) can move to the goal in the next step. Agent \(a_3\), whose priority is now higher than that of \(a_2\), then requests the node below and executes a \(90^\circ\) counterclockwise rotation. Finally, agent \(a_2\), which has already reached its goal node and whose priority is reset to its initial value, requests its current node and waits in place.

From the fourth timestep onward, both \(a_1\) and \(a_2\) remain at their goal nodes, while \(a_3\) requires one additional step to reach its target. The instance is solved at the fifth timestep.

\subsection{Cycle handling approach} \label{Sec. Cycle}

\begin{figure}[tb]
	\centering
	\includegraphics{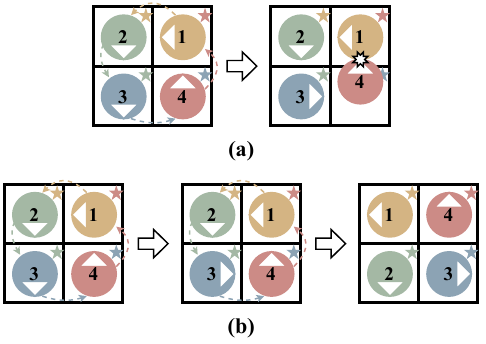}
	\caption{(a) An instance of cycle. (b) Cycle is handled by proposed approach.}
	\label{fig_cyc}
\end{figure}

When turn actions are incorporated, the node request mechanism inherited from PIBT may fail in the presence of circular request chains. Figure~\ref{fig_cyc}(a) illustrates such a case, where agents~1, 2, 3, and~4 have goal nodes A, B, C, and~D, respectively. Following the node request mechanism described in Section~\ref{sec:PIBTT*}, all agents successfully request their goal nodes. However, agent~3 must first rotate in place to align with its requested direction. This turn-induced delay prevents agents~1 and~2 from moving as expected and ultimately leads to a vertex conflict between agents~4 and~1.

This example shows that, in MAPF\textsubscript{T}, a circular request chain may invalidate the conflict-resolution logic based on recursive backtracking. We refer to such a request structure as a \textit{cycle} and formalize it as follows.

\begin{figure}[tb]
	\centering
	\includegraphics[width=0.7\columnwidth]{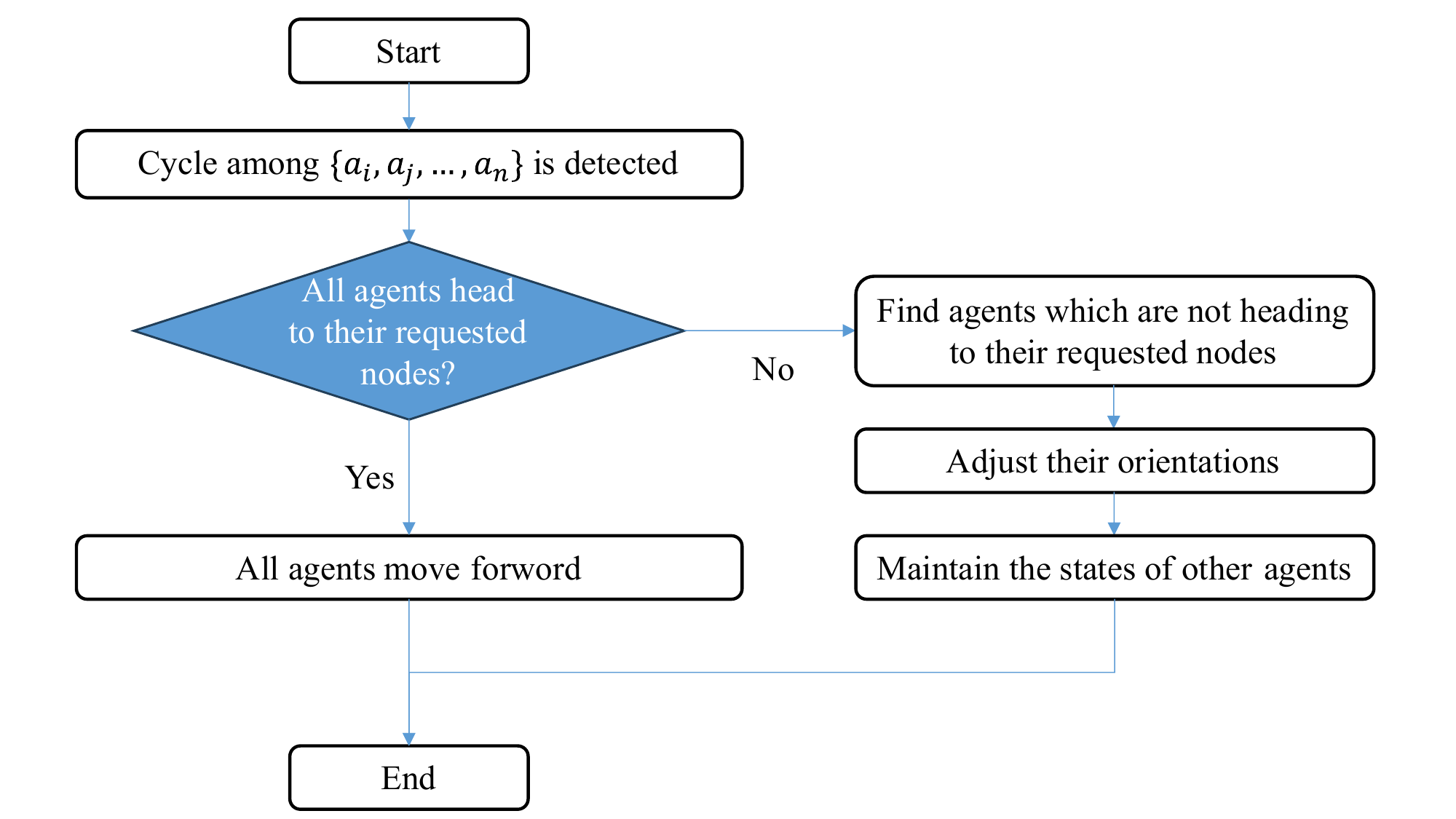}
	\caption{Cycle handling approach}
	\label{fig_cyc_ha}
\end{figure}

\begin{definition}
Consider a sequence of agents \(\mathcal{C} = \langle a_1, a_2, \ldots, a_n \rangle\) (\(n>2\)) involved in a recursive priority inheritance chain at timestep \(t\). Let \(\pi_i[t]\) denote the current node occupied by agent \(a_i\), and let \(v_i[t+1]\) denote the target node requested by \(a_i\). We define \(\mathcal{C}\) as a \textit{cycle} if and only if:
\begin{enumerate}
    \item For all \(1 \le i < n\), agent \(a_i\) requests the node currently occupied by its successor \(a_{i+1}\), i.e., \(v_i[t+1] = \pi_{i+1}[t]\);
    \item The last agent \(a_n\) requests the node occupied by the first agent \(a_1\), i.e., \(v_n[t+1] = \pi_1[t]\).
\end{enumerate}
\end{definition}

The next proposition identifies where a collision may arise once such a cycle is formed.

\begin{proposition}
Let \(\mathcal{C} = \langle a_1, a_2, \ldots, a_n \rangle\) be a cycle formed during recursive priority inheritance at timestep \(t\). The only potential collision induced by this cycle can occur between the terminator agent \(a_n\) and the initiator agent \(a_1\). In particular, no collision can occur between any intermediate pair \((a_i, a_{i+1})\) for \(1 \le i < n\).
\end{proposition}

\begin{proof}
In the recursive planning framework, the recursion unfolds along the request chain until it reaches the terminator agent \(a_n\), which determines its next-step action first and passes the resulting occupation status back up the chain. Consider any intermediate pair \((a_i, a_{i+1})\) with \(1 \le i < n\). Since agent \(a_{i+1}\) determines its action before \(a_i\), agent \(a_i\) can evaluate whether \(\pi_{i+1}[t]\) will be vacated at timestep \(t+1\). Under the MAPF\textsubscript{T} action model, an agent may either move forward, rotate \(90^\circ\) in place, or wait in place. If \(a_{i+1}\) is not aligned with its requested node, then its next-step action is an in-place rotation, and therefore \(\pi_{i+1}[t]\) remains occupied at timestep \(t+1\). This information is available to \(a_i\) through recursive backtracking. Hence, even if \(a_i\) is aligned with its requested node, it updates its action to wait in place once it detects that the node will remain occupied at timestep \(t+1\). Therefore, no collision can occur between any intermediate pair \((a_i, a_{i+1})\). The above safeguard does not necessarily apply to the pair \((a_n,a_1)\). When \(a_n\) determines its action, agent \(a_1\) is still in the recursion stack and its next-step state has not yet been finalized. Consequently, \(a_n\) cannot know whether \(\pi_1[t]\) will actually be vacated at timestep \(t+1\). If \(a_n\) is aligned with \(\pi_1[t]\) and chooses to move forward, while \(a_1\) is subsequently forced to rotate or wait in place, then a collision may occur. Therefore, among the conflicts induced by the cycle, the only unresolved one can occur between \(a_n\) and \(a_1\).
\end{proof}

To eliminate the vulnerability identified in Proposition~1, a naive conservative rule would simply block the terminator agent whenever the initiator’s next position is uncertain. While safe, such a rule may unnecessarily freeze the whole cycle and delay progress. Instead, we propose the cycle handling approach shown in Figure~\ref{fig_cyc_ha}, which preserves safety while allowing the cycle to make progress whenever possible. Once a cycle \(\mathcal{C}\) is detected, the algorithm first checks whether all agents in \(\mathcal{C}\) are already aligned with their requested directions. If so, all of them move simultaneously to their requested nodes. Otherwise, in the next timestep, agents in \(\mathcal{C}\) that are not aligned perform turn actions, while the remaining agents in \(\mathcal{C}\) wait in place, as illustrated in Figure~\ref{fig_cyc}(b).

\begin{proposition}
For any detected cycle \(\mathcal{C}\), the proposed cycle handling approach guarantees a conflict-free next-step update for all agents in \(\mathcal{C}\).
\end{proposition}

\begin{proof}
We consider the two cases of the cycle handling approach:

\begin{enumerate}
    \item \textit{All agents in \(\mathcal{C}\) are aligned.}  
    In this case, every agent in \(\mathcal{C}\) moves forward to its requested node. Since \(\mathcal{C}\) is a cycle, each requested node is exactly the current node of another agent in \(\mathcal{C}\), and each such node is vacated simultaneously by that agent. Therefore, no two agents move into the same node, and no agent moves into a node that remains occupied. Moreover, the motions follow the directed cycle consistently, so no edge conflict occurs. Hence, the simultaneous update is conflict-free.

    \item \textit{At least one agent in \(\mathcal{C}\) is not aligned.}  
    In this case, every misaligned agent performs an in-place turn, while every other agent in \(\mathcal{C}\) waits in place. Thus, no agent in \(\mathcal{C}\) attempts to enter a different node at timestep \(t+1\). Since the configuration at timestep \(t\) is collision-free, keeping all agents in \(\mathcal{C}\) on their current nodes preserves collision-freeness at timestep \(t+1\). Therefore, the next-step update is also conflict-free.
\end{enumerate}

Since both cases yield a conflict-free next-step update for all agents in \(\mathcal{C}\), the proposed cycle handling approach is correct. 
\end{proof}

\subsection{Node selection strategy}
\label{Sec. Node_Selection}
In PIBT-based algorithms, the node selection strategy (line \ref{Alg1, node_selection}, Algorithm \ref{algorithm}) determines the next preferred position for each agent. The node selection strategy plays a key role in the overall performance of the algorithm. An effective strategy can resolve deadlocks among agents, guide them to their goals efficiently, and consequently enhance the operational efficiency of the entire system.

\textbf{Local greedy strategy}. 
In the original PIBT algorithm, candidate nodes are ordered by their distances to the agent’s goal, where distances are computed on the static map (ignoring other agents) and can thus be preprocessed using A$^*$. However, under MAPF\textsubscript{T}, turn actions introduce additional costs that invalidate such heuristics: the cost of reaching the goal through an intermediate node depends not only on spatial distance but also on orientation transitions. This makes classical distance heuristics inaccurate for ranking neighboring nodes.

To address this challenge, we introduce a Local Greedy Strategy (LGS) that sorts candidate nodes according to the following cost function:
\begin{equation}
    TC(s,v_i,g)=CC(s,v_i)+RC(s',g),
\end{equation}
where \(TC(s,v_i,g)\) represents the total cost of reaching the goal node \(g\) from the current state \(s\) via a neighboring node \(v_i\). The term \(CC(s,v_i)\) denotes the cost from \(s\) to the first state \(s'\) upon arriving at \(v_i\), while \(RC(s',g)\) denotes the remaining cost from \(s'\) to \(g\). We next show that, in the single-agent setting, this ranking rule is consistent with shortest-path optimality.

\begin{proposition}
Consider the single-agent MAPF\textsubscript{T} problem on the state graph, and let \(RC(s,g)\) denote the exact optimal remaining cost from state \(s\) to the goal \(g\). For any current state \(s\), selecting the node \(v_i\in Neighbor(\pi_i[t])\) that minimizes
\[
TC(s,v_i,g)=CC(s,v_i)+RC(s',g)
\]
yields an optimal next-step decision.
\end{proposition}

\begin{proof}
Consider the single-agent MAPF\textsubscript{T} problem as a shortest-path problem on the state graph, where each node represents an agent state. For any state \(s\) that is not yet at the goal, every feasible path to the goal must first proceed toward one of the neighboring nodes of the current position. Let \(A\) denote the set of first states reached upon arriving at these neighboring nodes. Since every path from \(s\) to \(g\) must pass through some state in \(A\), the optimal cost from \(s\) to \(g\) satisfies
\[
d(s,g)=\min_{a\in A}\{d(s,a)+d(a,g)\}.
\]
For each neighboring node \(v_i\), the term \(CC(s,v_i)\) is exactly the cost of reaching the corresponding first arrival state \(s'\), and \(RC(s',g)\) is the exact optimal remaining cost from \(s'\) to \(g\). Therefore, minimizing \(TC(s,v_i,g)\) is equivalent to choosing the neighboring node whose associated first-arrival state lies on a shortest path from \(s\) to \(g\). Hence, the selected \(v_i\) yields an optimal next-step decision.
\end{proof}

While \(CC\) is straightforward to compute, evaluating \(RC\) requires solving a shortest-path problem in the state space. To avoid repeated online searches, we introduce a preprocessing step during initialization. Specifically, for each agent, we perform a breadth-first search on the state graph to compute the optimal costs from all reachable states to its goal. Since each physical node has four possible orientation states, this procedure generates \(4|A|\) cost tables in total, corresponding to the four goal-oriented states of all agents. These tables are stored offline and can be queried in constant time during node selection, thereby significantly improving planning efficiency.

\begin{figure}[t]
	\centering
	\includegraphics{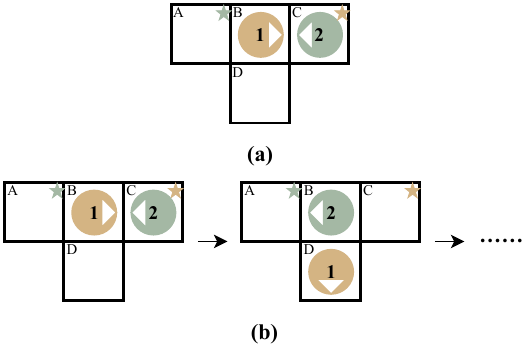}
	\caption{Deadlock in PIBT and swap process}
	\label{fig_deadlock}
\end{figure}

\textbf{Swap pattern detector}. The swap pattern detector is designed to resolve deadlock and livelock situations that may arise in PIBT-based planning. In the original PIBT, all agents are guaranteed to reach their goal nodes at least once within finite time under the condition that every pair of adjacent nodes belongs to a simple cycle. When this condition is violated, agents may stop moving after a certain timestep, resulting in deadlock. As illustrated in Figure~\ref{fig_deadlock}(a), a higher-priority agent $a_1$ requests node $C$, currently occupied by $a_2$. The recursive PIBT$^{\text{*}}_{\text{T}}$ procedure for $a_2$ fails because moving to $B$ or staying at $C$ would conflict with $a_1$. Consequently, $a_1$ is forced to remain at $B$ according to the LGS, leading to a state of mutual obstruction. Even when the condition is satisfied, agents may alternately reach their goals due to priority resets without making simultaneous progress, leading to livelock.

Most deadlock and livelock cases stem from local mutual blocking between two agents and can be resolved through a simple cooperative swap, rather than strictly following the LGS. Motivated by this observation, we introduce a swap pattern detector (SPD) inspired by \cite{okumura2023improving}, which identifies situations where a higher-priority agent should temporarily yield to enable such cooperation (Algorithm~\ref{algorithm_swap}).

\begin{algorithm}[!b]
    \caption{Swap pattern detector for agent $a_i$}
    \LinesNumbered
    \label{algorithm_swap}
    \SetKwFunction{SwapFunction}{SwapAgent}
    \SetKwProg{Fn}{function}{:}{end function}
    \Fn{\SwapFunction{$a_i,P_i,\pi_i[t]$}}{
        \If{$P_i[0]$\textnormal{ is occupied by }$a_j$}{ \label{Alg2, start_usual_swap}
            %\textnormal{Check whether swap is required and possible by a 2-stage simulation}\;
            \If{\textnormal{Swap is required and possible}}{
                \KwRet{$a_j$}\;
            }
        } \label{Alg2, end_usual_swap}
        \ForEach{$u\in \pi_i[t].neighbor$}{ \label{Alg2, start_clear_operation}
            \If{$u$\textnormal{ is occupied by }$a_k$\textnormal{ and }$u\neq P_i[0]$}{ 
                %\textnormal{Check whether swap is required and possible by a 2-stage simulation}\;
                \If{\textnormal{Swap is required and possible}}{
                    \KwRet{$a_k$}\;
                }
            }
        } \label{Alg2, end_clear_operation}
    }

    \BlankLine
    
    \If{$a_j\leftarrow $ \SwapFunction{$a_i,P_i,\pi_i[t]$}$\ne$\textnormal{NULL}}{ \label{Alg2, swap_agent}
        $P_i$.reverse()\; \label{Alg2, reverse}
        Compute and update action for $a_j$ to move to $\pi_i[t]$ after $\pi_i[t+1]$ has been planned.\;
    }
\end{algorithm}

After obtaining the preference list $P_i$ for agent $a_i$, the detector first checks whether the most preferred node $P_i[0]$ is occupied by another agent $a_j$. If so, a two-stage simulation is performed to determine whether a swap is both \textit{required} and \textit{possible}. The first stage evaluates swap necessity by simulating $a_i$ pushing toward $a_j$ while $a_j$ is displaced to adjacent nodes. A swap is deemed required if (i) $a_j$ is forced into a dead-end corridor and the relative goal distances indicate a “crossing” situation, i.e., $a_i$ is closer to $g_j$ than $a_j$ is, and $a_j$ is closer to $g_i$ than $a_i$ is.  (ii) $a_i$ reaches its goal but blocks $a_j$’s only path forward. In both cases, strictly following the LGS would trap $a_j$, whereas a swap allows both agents to progress. If $a_j$ instead reaches a node with sufficient maneuvering space (degree more than 2), the swap is considered unnecessary.

The second stage checks whether the swap is possible by reversing the roles: $a_j$ iteratively pushes $a_i$ toward adjacent nodes. If $a_i$ reaches a dead end (degree 1), the swap is infeasible. If $a_i$ reaches a node of degree more than 2, the swap is possible. When a swap is confirmed, $a_j$ is marked as the swap agent and the preference list $P_i$ is reversed (line~\ref{Alg2, swap_agent}), allowing $a_i$ to concede temporarily.

Beyond the primary case, Algorithm~\ref{algorithm_swap} also examines whether nearby agents occupying non-preferred neighboring nodes should enter a dead-end corridor (lines~\ref{Alg2, start_clear_operation}--\ref{Alg2, end_clear_operation}), extending swap detection to a broader set of local configurations. The same two-stage simulation logic is applied in this case.

Figure \ref{fig_deadlock}(b) illustrates the resolution of the deadlock instance through the proposed SPD. When $a_1$ identifies that its preferred node $C$ is occupied by $a_2$, it initiates the two-stage simulation. The first stage confirms the necessity of a swap because $a_2$ is restricted to a degree-1 node. The second stage verifies the feasibility of the procedure because $a_1$ occupies node $B$, which has a degree more than 2 and provides sufficient space to reposition. Following these results, $a_1$ reverses its preference list $P_1$ and moves toward node $D$ instead of $C$. By selecting this originally least-preferred candidate, $a_1$ effectively vacates the path and guides $a_2$ toward node $B$, thereby resolving the deadlock.

\begin{figure}[t]
	\centering
	\includegraphics{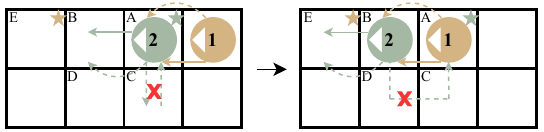}
	\caption{Agent at its goal is continuously pushed away.}
	\label{fig_escape}
\end{figure}

\textbf{\textit{k}-Push escape trigger}.
The short-sighted nature of the LGS can significantly degrade solution quality in some cases. As illustrated in the first configuration of Figure~\ref{fig_escape}, agent~1 requests node~A, which is the goal of agent~2, to move toward its own goal~E. This triggers priority inheritance, forcing agent~2, already located at its goal, to vacate node~A by choosing between nodes~B and~C. Since $TC(s_2,v_B,g_2)=1+3=4$ and $TC(s_2,v_C,g_2)=2+3=5$, agent~2 is pushed to node~B. In the next timestep (the second configuration in Figure~\ref{fig_escape}), the same decision pushes agent~2 further to node~E. As a result, agent~2 is repeatedly pushed away until agent~1 reaches its goal, instead of taking a more efficient alternative path back to node~A, leading to an unnecessarily long makespan. 

In more severe cases, particularly when the SPD module fails, two agents may repeatedly push each other without making effective progress toward their goals, resulting in a livelock. This issue arises when a low-priority agent is repeatedly invoked by the same higher-priority agent without escaping the pushing pattern. The root cause lies in the short-sighted and memory-less nature of the decision process.

To address this problem, we introduce a simple escape mechanism that detects repeated priority inheritance between the same pair of agents and forces exploration once such a pattern persists. Specifically, we maintain a push-time table $PT$ of size $|A|^2$, which records how many times each agent has been invoked for priority inheritance by another specific agent. All entries in $PT$ are initialized to zero. Whenever agent $a_j$ is recursively invoked by a higher-priority agent $a_i$ via the PIBT$^{*}_{\text{T}}$ procedure and eventually moves to a new node, $PT(a_j,a_i)$ is increased by one. Once this value reaches a predefined threshold $k$, the next time $a_j$ is pushed by $a_i$, its node preference list $P_j$ is randomly shuffled to encourage exploration of alternative routes and break the repetitive pushing pattern.

The choice of $k$ plays an important role in performance. A small $k$ may cause agents to frequently deviate from their individually optimal paths in high-density scenarios, degrading solution quality. Conversely, a large $k$ allows repetitive pushing to persist for many timesteps, limiting the improvement. In the next section, a suitable value of $k$ is determined through numerical experiments.

\subsection{Complexity analysis}
As shown in Algorithm \ref{algorithm}, within a single timestep of planning, sorting the agents based on updated priorities (line 2) requires $O(|A|lg|A|)$. During the planning process for each agent, iterating over the candidate nodes requires $O(\Delta(G)+1)$, where $O(\Delta(G))$ is the maximum degree of the graph. Both detecting cycles and handling cycles requires $O(|C|)$, where $|C|$ is the number of agents in the cycle with $|C|\le|A|$, resulting in an overall complexity of $O(|A|)$. Assuming other operations are completed in constant time $O(K)$, the proposed algorithm achieves a time complexity of $O(|A|\cdot(lg|A|+|A|\Delta(G)+K))$. It is evident that the proposed method requires only polynomial time complexity for planning within a single timestep, offering a significant computational advantage over algorithm with exponential time complexity \citep{zhang2023efficient}.

The time complexity for creating the cost table during initialization is $O(4|A|\cdot(4|V|\cdot3))=O(|A|\cdot|V|)$. In large-scale instances involving extensive maps or a high number of agents, the preprocessing time will increase accordingly. However, experiments on benchmark maps have demonstrated acceptable preprocessing speeds, which will be further explained in the numerical experiments section.

\section{Numerical results} \label{sec:experiment}

This section evaluates the effectiveness of the proposed mechanisms and the overall performance of PIBT$^{\text{*}}_{\text{T}}$. The algorithm is implemented in C++ by extending the public implementations of \citet{okumura2019priority} and \citet{okumura2022priority}. 
The source code are publicly available on GitHub.\footnote{\url{https://github.com/tzy82065/PIBTT2}}. Supplementary videos are provided to illustrate the simulation and real-robot demonstrations. All experiments are executed on a machine equipped with an Intel i5-12500 CPU (3.00\,GHz) and 16\,GB of RAM.

\subsection{Experimental setup}

We evaluate the proposed algorithm through six sets of experiments, covering parameter calibration, mechanism analysis, benchmark comparison, scalability testing, and realistic execution validation. The first set investigates the impact of the parameter \(k\) in the \(k\)-Push Escape Trigger (PET) on solution quality. Based on the calibrated value of \(k\), the second set conducts an ablation study on PET and SPD, evaluating their effects on success rate and solution quality. The third set compares the proposed PIBT$^{\text{*}}_{\text{T}}$ algorithm with two baseline methods in terms of success rate, computational efficiency, and solution quality.

The fourth set performs a large-scale stress test to assess the scalability of PIBT$^{\text{*}}_{\text{T}}$ in terms of success rate and runtime. The fifth set compares the proposed MAPF\textsubscript{T} solver with the classical MAPF solver PIBT, aiming to quantify the gap between theoretical planning cost and actual execution cost when the generated paths are executed by robots with realistic kinematic constraints. Finally, the sixth set presents a case study on a typical manufacturing system to evaluate the practical advantage of the proposed algorithm in an industrially relevant scenario.

\textbf{Benchmarks and Instances.} %The first five sets of experiments are conducted on five benchmark maps from the MovingAI repository\footnote{\url{https://movingai.com/benchmarks/mapf/index.html}\label{MAPF_Benchmark}}: \textit{empty-32-32}, \textit{random-32-32-20}, \textit{room-64-64-8}, \textit{warehouse-10-20-10-2-2}, and \textit{den520d}. For each map and each agent number (varying from 5 to 100), we utilize 25 distinct random scenarios provided by the benchmark\textsuperscript{\ref{MAPF_Benchmark}} with varying starts and goal positions. To guarantee a fair comparison, all competing methods are evaluated on the exact same set of instances. The stress test is performed specifically on the \textit{warehouse} map. An execution is terminated and marked as a failure if the runtime exceeds 60 seconds \citep{zhang2023efficient}.
The benchmark experiments are conducted on five standard maps from the MovingAI repository\footnote{\url{https://movingai.com/benchmarks/mapf/index.html}\label{MAPF_Benchmark}}: \textit{empty-32-32}, \textit{random-32-32-20}, \textit{room-64-64-8}, \textit{warehouse-10-20-10-2-2}, and \textit{den520d}. For each map and each agent number, ranging from 5 to 100 with an interval of 5, we use 25 random scenarios provided by the benchmark\textsuperscript{\ref{MAPF_Benchmark}}, with different start and goal locations. To ensure a fair comparison, all competing methods are tested on identical start--goal scenarios for each map and agent number. The large-scale stress test is conducted on the \textit{warehouse-10-20-10-2-2} map. Following the termination criterion used by \citet{zhang2023efficient}, a run is counted as successful only if a conflict-free solution is found within 60 seconds; otherwise, it is terminated and marked as a failure.

\textbf{Baselines.}
The baselines include PIBT$_{\text{T}}$ \citep{tao2025fast}, an earlier priority-inheritance-based MAPF\textsubscript{T} solver, and CBS$_{\text{T}}$ \citep{zhang2023efficient,li2021pairwise}, a sum-of-cost-optimal MAPF\textsubscript{T} solver. These two methods are selected because, to the best of our knowledge, they are the only publicly available solvers that adopt the same MAPF\textsubscript{T} action model as the proposed method.

\subsection{Effect of parameter k in PET}

This subsection investigates the effect of the threshold parameter \(k\) in PET. Different values of \(k\) are tested on the benchmark maps to examine how the escape frequency affects solution quality under different map structures and agent densities. We report the makespan gap, defined as the relative increase in makespan compared with the best makespan obtained among all tested \(k\) values for the same map and agent number. Table~\ref{tab_k} summarizes the results for the 100-agent cases.

\begin{table}[t]
	%\caption{Makespan gap with different $k$ ($|A|=100$)}
    \caption{Makespan gaps under different PET threshold values \(k\) for 100-agent instances.}
	\centering
	\begin{tabular}{ccccccc}
		\toprule[2pt]
		Map & $k=1$ & $k=2$ & $k=3$ & $k=4$ & $k=5$ & $k=\infty$ \\ % 增加表头
		\midrule
		$Empty$ & 0 & 3.25\% & 4.42\% & 8.08\% & 10.50\% & 29.42\% \\
		$Random$ & 4.63\% & 0 & 0.84\% & 2.43\% & 2.99\% & 4.19\% \\
		$Room$ & 6.59\% & 0.64\% & 0 & 2.61\% & 0.86\% & 1.41\% \\
		$Warehouse$ & 0 & 0.02\% & 0.40\% & 0.25\% & 1.94\% & 36.26\% \\
		$Den520d$ & 0 & 0.11\% & 0.11\% & 0 & 0.11\% & 2.27\% \\
		\bottomrule[1pt]
	\end{tabular}
	\label{tab_k}
\end{table}

The results show that the choice of \(k\) affects solution quality differently across maps. On the smaller and denser maps, namely \textit{empty}, \textit{random}, and \textit{room}, priority inheritance occurs more frequently, making the escape frequency more influential. A very small \(k\) triggers exploration too aggressively and may cause agents to deviate repeatedly from individually efficient paths. In contrast, a large \(k\) allows inefficient pushing patterns to persist for more timesteps before intervention, reducing the effectiveness of PET. As a result, intermediate values of \(k\) often provide a better balance between exploration and path efficiency, although the best value differs across maps.

For the larger maps, \textit{warehouse} and \textit{den520d}, the lower effective agent density reduces the frequency of repeated priority inheritance. Consequently, the impact of \(k\) is less pronounced, and small values of \(k\) generally perform well. The case \(k=\infty\), which corresponds to disabling the escape trigger, leads to a clear degradation on several maps, especially \textit{empty} and \textit{warehouse}. This confirms that PET is useful for mitigating inefficient repeated pushing. Considering the overall robustness across different maps, we set \(k=2\) in the following experiments.

After fixing \(k=2\), we further assess the stability of the proposed algorithm under the stochastic random-shuffling mechanism introduced by the PET module. For each map, we select one 100-agent instance and execute the algorithm with 10 different random seeds. Table~\ref{tab_variability} reports the mean, standard deviation, and coefficient of variation (CV) of both sum of costs and makespan. 

The results show that the stochasticity introduced by PET has different effects across map structures. On the unstructured \textit{random} map, the variability is relatively higher, with a CV of 5.13\% for sum of costs and 13.08\% for makespan. This higher variability can be explained by the irregular obstacle layout of the \textit{random} map, where local greedy decisions are more likely to generate repeated pushing patterns and local blocking configurations \citep{tao2025fast}. As a result, the PET module is triggered more frequently, and its random shuffling mechanism may lead agents to choose different escape routes under different random seeds. These different local choices can further propagate through priority inheritance and produce noticeable variations in solution quality, especially in makespan. In contrast, on structured maps such as \textit{warehouse} and \textit{den520d}, the CV remains very low, indicating that the proposed method is more stable in regular environments. Overall, the variability analysis suggests that PET improves the algorithm performance while maintaining acceptable stability.

%In contrast, on structured maps such as \textit{warehouse} and \textit{den520d}, the CV remains very low, indicating that the proposed method is highly stable in regular environments. Overall, the variability analysis suggests that PET improves the robustness of the search behavior while maintaining acceptable reproducibility.

\subsection{Ablation experiment for PET and SPD}

\begin{table}[]
    \caption{Variability of solution quality with different random seeds ($|A|=100$, 10 runs)}
    \centering
    \begin{tabular}{ccccc}
        \toprule[2pt]
        \multirow{2}{*}{\textbf{Map}} & \multicolumn{2}{c}{\textbf{Sum of Costs}} & \multicolumn{2}{c}{\textbf{Makespan}} \\
        \cmidrule(lr){2-3} \cmidrule(lr){4-5}
         & Mean $\pm$ SD & CV & Mean $\pm$ SD & CV \\
        \midrule
        $Empty$ & $3386.9 \pm 82.0$ & 2.42\% & $57.2 \pm 3.9$ & 6.74\% \\
        $Random$ & $5891.4 \pm 302.5$ & 5.13\% & $116.9 \pm 15.3$ & 13.08\% \\
        $Room$ & $10175.5 \pm 242.9$ & 2.39\% & $166.9 \pm 7.6$ & 4.55\% \\
        $Warehouse$ & $11539.7 \pm 20.1$ & 0.17\% & $211.3 \pm 0.7$ & 0.32\% \\
        $Den520d$ & $17685.7 \pm 67.5$ & 0.38\% & $402.2 \pm 1.0$ & 0.26\% \\
        \bottomrule[1pt]
    \end{tabular}
    \label{tab_variability}
\end{table}

% \begin{figure}[tb]
% 	\centering
% 	\includegraphics[width=0.7\linewidth]{fig/ablation.pdf}
% 	\caption{Ablation experiment of PET and SPD.}
% 	\label{fig_ablation}
% \end{figure}

\begin{figure}[tb]
	\centering
	\includegraphics[width=\linewidth]{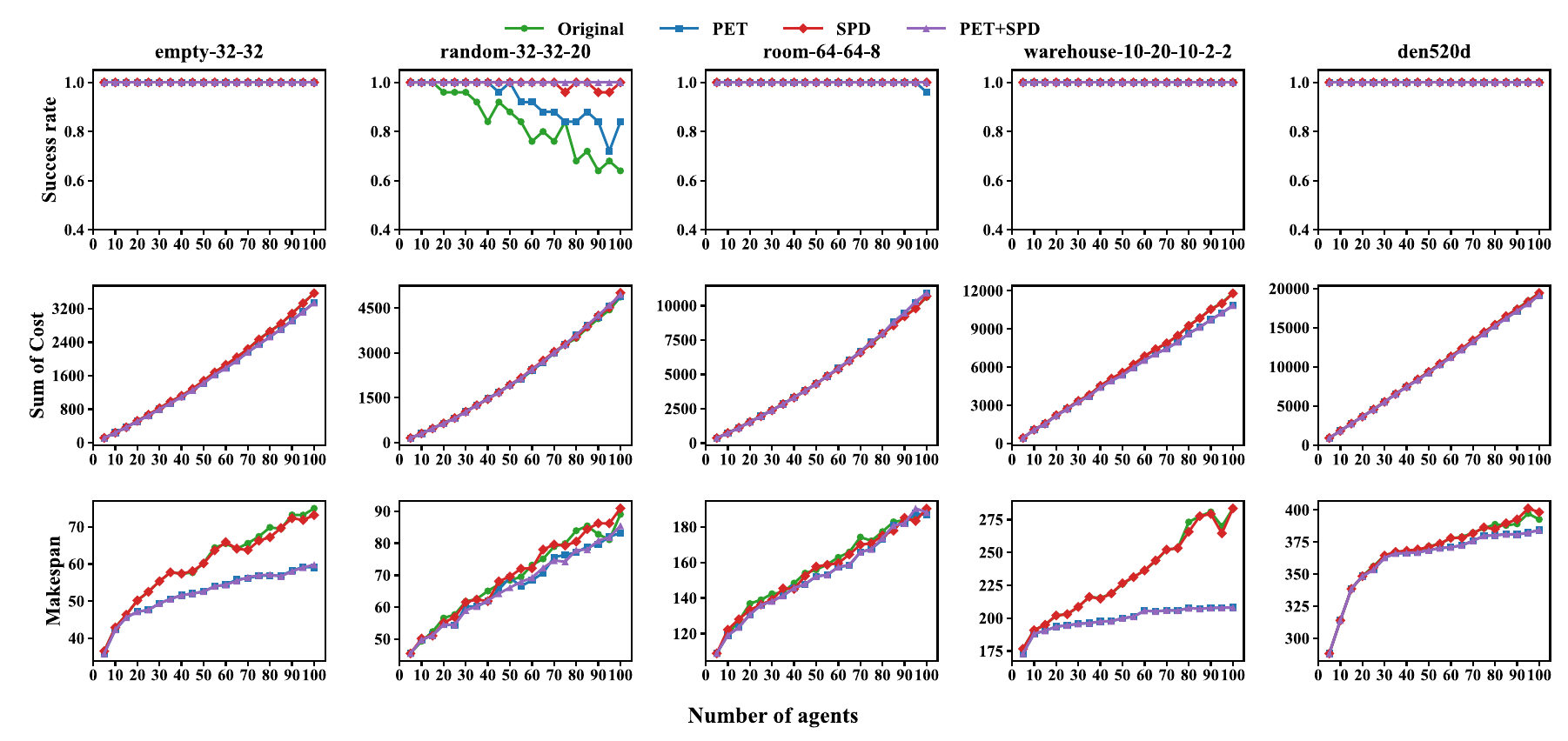}
	%\caption{Ablation experiment of PET and SPD.}
    \caption{Ablation study of the PET and SPD modules on benchmark maps under different numbers of agents.}
	\label{fig_ablation}
\end{figure}

This subsection evaluates the individual and combined contributions of the SPD and PET modules. Starting from the original PIBT$_{\text{T}}$ algorithm without either module, we progressively add PET, SPD, and their combination, resulting in four variants: Original, PET, SPD, and PET+SPD. These variants are evaluated on the same test instances, with the number of agents ranging from 5 to 100.

Figure~\ref{fig_ablation} presents the ablation results. We first focus on the success rates of the four variants. PIBT$_{\text{T}}$ (Original) exhibits a relatively low success rate on the \textit{random} map, and its success rate decreases rapidly as the number of agents increases. The failures can be attributed to two factors. First, maps such as \textit{random-32-32-20} often violate the graph condition required by PIBT-style algorithms to generate feasible solutions, namely that every pair of adjacent nodes should belong to a simple cycle \citep{okumura2022priority}. When this condition is not satisfied, agents may become permanently blocked, resulting in deadlock. Second, even when such graph condition holds, PIBT$_{\text{T}}$ does not guarantee that all agents can reach and remain at their goal nodes simultaneously. Agents may repeatedly arrive at and leave their goals without ever forming a terminal configuration in which all agents are simultaneously located at their goals, resulting in livelock. Introducing PET alone resolves some livelock cases, but yields only limited improvement in success rate. By contrast, SPD addresses most local deadlocks, although some livelock cases remain unresolved. Combining PET and SPD enables all test instances to be successfully solved.

We then analyze the solution-quality metrics of the four variants. The SoC values are relatively close across variants, whereas the differences in makespan are more pronounced, especially on the \textit{empty} and \textit{warehouse} maps. As shown in Figure~\ref{fig_ablation}, PIBT$_{\text{T}}$ (Original) performs relatively poorly on these two maps. Adding SPD alone brings limited benefit in reducing the makespan, although it substantially improves the success rate across the benchmark maps. In contrast, incorporating PET leads to clear makespan reductions. This improvement is particularly evident in structured environments such as the \textit{warehouse} map, where the regular obstacle layout often causes agents to form queues and engage in repeated pushing patterns. PET disrupts such inefficient local behaviors by encouraging agents to explore alternative routes, thereby reducing congestion and improving the overall makespan. In summary, SPD mainly improves the success rate by resolving local deadlocks, whereas PET primarily improves solution quality by mitigating livelock and repeated pushing behaviors.

\subsection{Overall performance comparison}

\begin{figure*}[t]
	\centering
	\includegraphics[width=\linewidth]{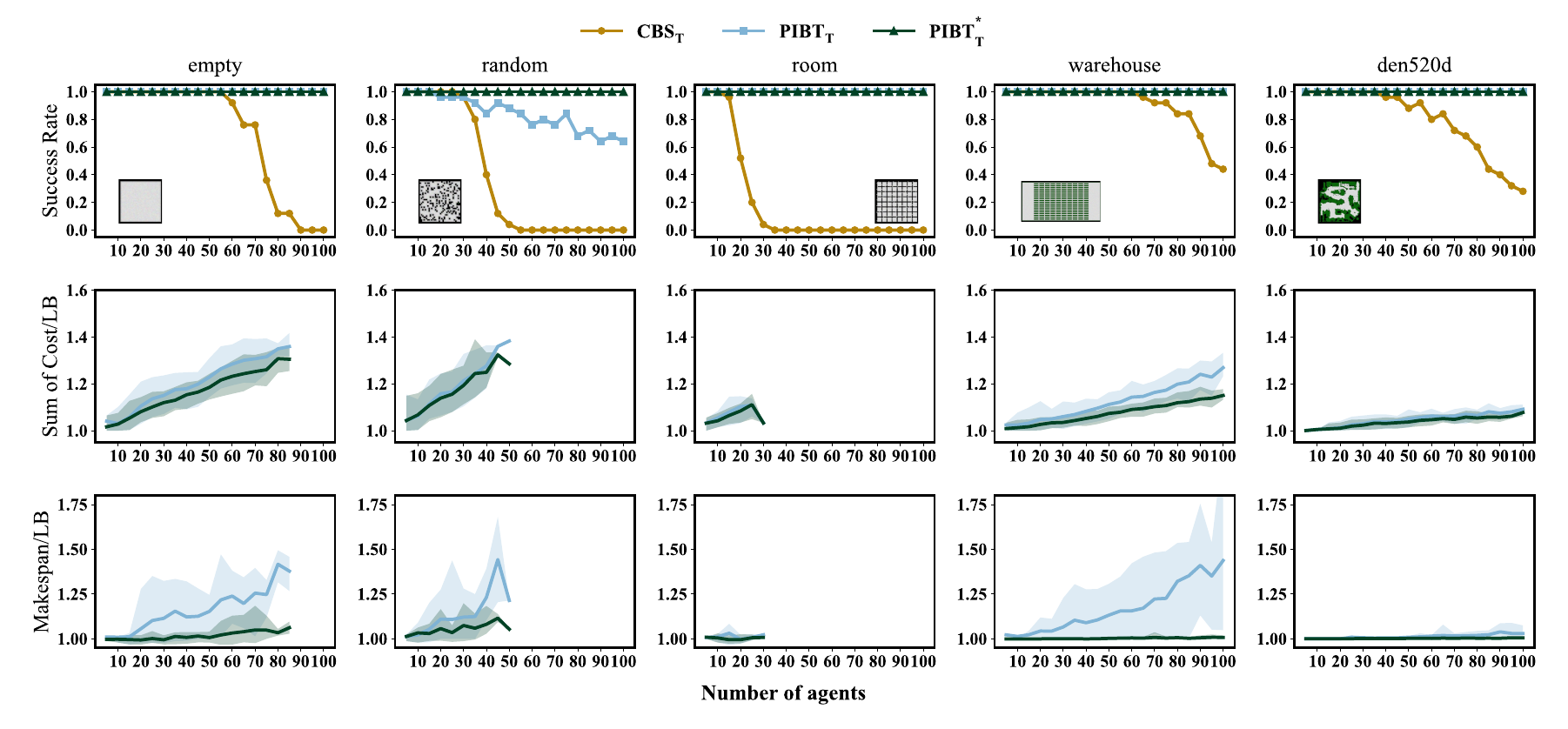}
	%\caption{Success rate of CBS$_{\text{T}}$, PIBT$_{\text{T}}$, and PIBT$^{\text{*}}_{\text{T}}$, and cost and makespan gaps of
%PIBT$_{\text{T}}$ and PIBT$^{\text{*}}_{\text{T}}$ relative to CBS$_{\text{T}}$ (LB).}
    \caption{Success rates of CBS$_{\text{T}}$, PIBT$_{\text{T}}$, and PIBT$^{\text{*}}_{\text{T}}$, together with the SoC and makespan ratios of PIBT$_{\text{T}}$ and PIBT$^{\text{*}}_{\text{T}}$ relative to the CBS$_{\text{T}}$ reference solutions.}
	\label{fig_subplot}
\end{figure*} 

\begin{table}[t]
\centering
\caption{Average runtime of PIBT$^{\text{*}}_{\text{T}}$ and CBS$_{\text{T}}$ on five benchmark maps (ms).}
\resizebox{0.7\columnwidth}{!}{
\large
\begin{tabular}{ccccccccc}
\hline
\toprule[2pt]
\multirow{2}{*}{\textbf{Map}}                & \multirow{2}{*}{\textbf{Method}} & \multicolumn{7}{c}{\textbf{Number of agents}}                                                         \\ \cline{3-9} 
                                             &                                  & \textbf{10} & \textbf{20}  & \textbf{30}  & \textbf{40}  & \textbf{50}  & \textbf{60}  & \textbf{70}  \\ \hline
\multirow{2}{*}{\textit{Empty}}     & \textbf{PIBT$^{\text{*}}_{\text{T}}$}                   & \textbf{4}  & \textbf{7}   & \textbf{10}  & \textbf{14}  & \textbf{17}  & \textbf{21}  & \textbf{25}  \\
                                             & CBS\textsubscript{T}                             & 1           & 5            & 15           & 63           & 533          & 4433         & 13244        \\
\multirow{2}{*}{\textit{Random}}    & \textbf{PIBT$^{\text{*}}_{\text{T}}$}                   & \textbf{4}  & \textbf{7}   & \textbf{10}  & \textbf{14}  & \textbf{17}  & \textbf{22}  & \textbf{26}  \\
                                             & CBS\textsubscript{T}                             & 4           & 44           & 3900         & 20464        & 32508        & -            & -            \\
\multirow{2}{*}{\textit{Room}}      & \textbf{PIBT$^{\text{*}}_{\text{T}}$}                   & \textbf{12} & \textbf{23}  & \textbf{33}  & \textbf{44}  & \textbf{59}  & \textbf{75}  & \textbf{83}  \\
                                             & CBS\textsubscript{T}                             & 884         & 9206         & 1361         & -            & -            & -            & -            \\
\multirow{2}{*}{\textit{Warehouse}} & \textbf{PIBT$^{\text{*}}_{\text{T}}$}                   & \textbf{28} & \textbf{54}  & \textbf{76}  & \textbf{100} & \textbf{123} & \textbf{147} & \textbf{170} \\
                                             & CBS\textsubscript{T}                             & 25          & 66           & 473          & 596          & 1810         & 6004         & 5788         \\
\multirow{2}{*}{\textit{Den520d}}   & \textbf{PIBT$^{\text{*}}_{\text{T}}$}                   & \textbf{90} & \textbf{166} & \textbf{247} & \textbf{315} & \textbf{389} & \textbf{463} & \textbf{548} \\
                                             & CBS\textsubscript{T}                             & 34          & 324          & 1776         & 2444         & 6862         & 13749        & 16266        \\
                                             \bottomrule[2pt]
\end{tabular}
}
\label{tab_runtime}
\end{table}

To evaluate the overall performance of PIBT$^{\text{*}}_{\text{T}}$, we compare it with two baseline algorithms, CBS$_{\text{T}}$ and PIBT$_{\text{T}}$, on the five benchmark maps. Figure~\ref{fig_subplot} reports the success rates of the three algorithms. As shown, PIBT$^{\text{*}}_{\text{T}}$ achieves a 100\% success rate across all tested instances, whereas the baseline algorithms experience increasing failure rates as the number of agents grows.

For CBS$_{\text{T}}$, the success rate declines more rapidly on the three smaller maps, namely \textit{empty}, \textit{random}, and \textit{room}, than on the larger maps, indicating its sensitivity to agent density. CBS$_{\text{T}}$ is a sum-of-cost-optimal solver based on the Conflict-Based Search framework. Although its low-level search is polynomial, the high-level Constraint Tree (CT) can grow exponentially in the worst case. Therefore, in dense scenarios with frequent conflicts, the CT may expand rapidly and cause the algorithm to exceed the 60-second runtime limit, as reported in Table~\ref{tab_runtime}.
%For CBS$_{\text{T}}$, the success rate declines rapidly on the first three smaller maps once the number of agents exceeds a critical threshold. Although it performs better on larger maps due to lower agent density, the success rate still decreases. CBS$_{\text{T}}$ is an sum-of-cost-optimal solver based on the Conflict-Based Search framework. While the low-level search is polynomial, the high-level Constraint Tree (CT) can grow exponentially in the worst case. In scenarios with high agent density and frequent conflicts, the CT expands explosively, causing the algorithm to exceed the 60-second runtime limit, as detailed in Table~\ref{tab_runtime}.
PIBT$_{\text{T}}$ exhibits a different failure pattern. It has polynomial-time single-step planning, but its success rate still drops significantly on the \textit{random} map due to deadlocks and livelocks \citep{tao2025fast}. PIBT$^{\text{*}}_{\text{T}}$ addresses these limitations by retaining the lightweight priority-inheritance mechanism while incorporating the SPD and PET modules. The SPD module resolves local deadlocks, whereas the PET module mitigates livelock and inefficient repeated pushing behaviors. Together, these mechanisms enable PIBT$^{\text{*}}_{\text{T}}$ to solve all tested instances.

%In contrast, PIBT$_{\text{T}}$ is a polynomial-time algorithm , yet its success rate also drops significantly due to livelocks and deadlocks \citep{tao2025fast}. Our proposed PIBT$^{\text{*}}_{\text{T}}$ overcomes these issues. It retains the low computational complexity of the priority inheritance mechanism but integrates the PET and SPD modules. These modules effectively resolve most deadlocks and livelocks, enabling PIBT$^{\text{*}}_{\text{T}}$ to solve all instances.

To evaluate the solution quality of PIBT$^{\text{*}}_{\text{T}}$ and PIBT$_{\text{T}}$, we further report the ratios of SoC and makespan relative to the corresponding CBS$_{\text{T}}$ reference values in Figure~\ref{fig_subplot}. For SoC, the reference value is the optimal objective value returned by CBS$_{\text{T}}$. For makespan, we use the makespan of the solution returned by CBS$_{\text{T}}$ as the reference value. Therefore, these ratio curves are plotted only for instances where CBS$_{\text{T}}$ successfully returns a solution within the 60-second time limit. The solid lines represent the mean ratios, and the shaded areas indicate the 95\% confidence intervals.

The results show that PIBT$^{\text{*}}_{\text{T}}$ consistently improves solution quality over PIBT$_{\text{T}}$. Across all benchmark maps, the ratios of PIBT$^{\text{*}}_{\text{T}}$ are lower than those of PIBT$_{\text{T}}$, especially in terms of makespan. As the number of agents increases, the quality gap between PIBT$^{\text{*}}_{\text{T}}$ and the CBS$_{\text{T}}$ reference generally becomes larger, particularly for SoC, because denser instances induce more agent interactions, detours, and priority-inheritance chains. Nevertheless, within the range where CBS$_{\text{T}}$ can solve the instances, PIBT$^{\text{*}}_{\text{T}}$ keeps the SoC and makespan ratios below 1.5 in all tested cases. This indicates that the proposed method maintains a practically acceptable quality gap while achieving substantially better scalability and success rate.

\begin{figure}[t]
	\centering
	\includegraphics[width=0.7\linewidth]{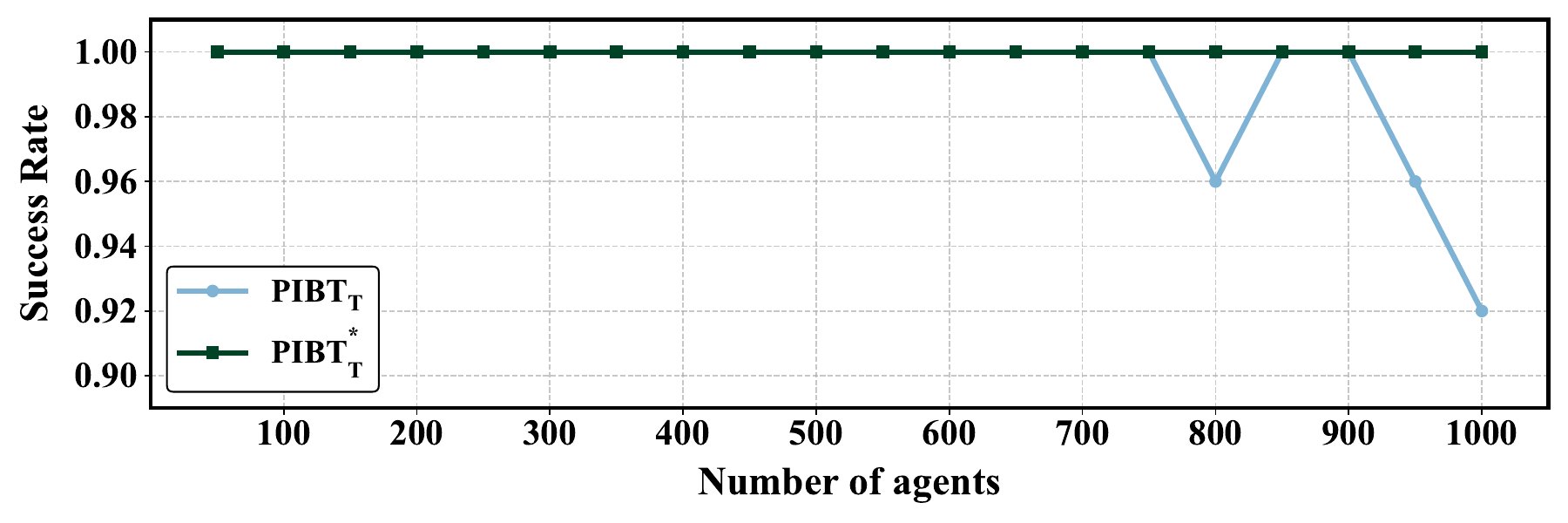}
	\caption{Success rate of PIBT$_{\text{T}}$ and PIBT$^{\text{*}}_{\text{T}}$ on the \textit{warehouse} map with up to 1{,}000 agents.}
	\label{fig_sr1k}
\end{figure}

\begin{figure}[t]
	\centering
	\includegraphics[width=0.7\linewidth]{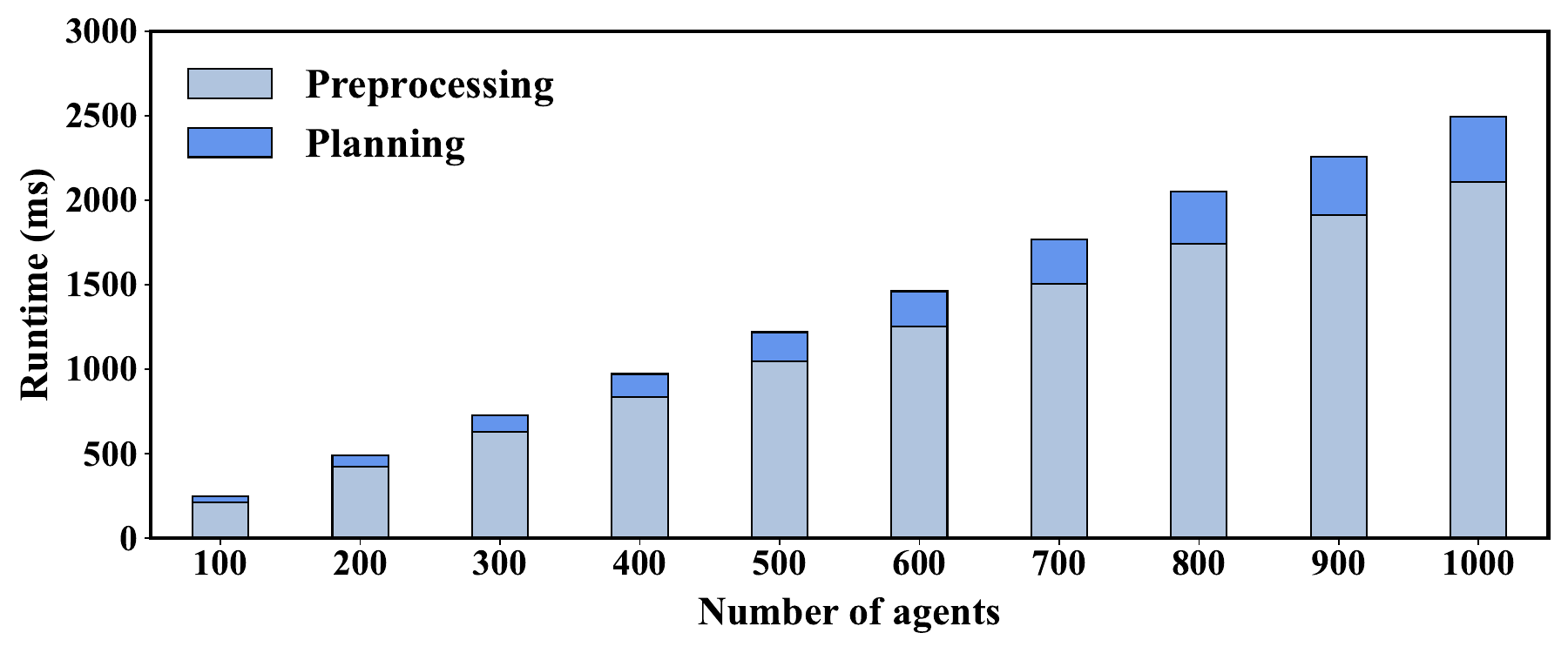}
	\caption{Runtime of PIBT$^{\text{*}}_{\text{T}}$ on the \textit{warehouse} map with up to 1{,}000 agents.}
	\label{fig_r1k}
\end{figure}

\subsection{Large-scale scenario with up to 1000 agents}
To test the scalability of the proposed method, we increase the number of agents from 50 to 1{,}000 on the \textit{warehouse} map. Figure~\ref{fig_sr1k} shows the success rates of PIBT$_{\text{T}}$ and PIBT$^{\text{*}}_{\text{T}}$. As shown, the success rate of PIBT$_{\text{T}}$ slightly decreases to 92\% when the number of agents reaches 1{,}000, whereas PIBT$^{\text{*}}_{\text{T}}$ successfully solves all tested instances.

We further analyze the runtime of PIBT$^{\text{*}}_{\text{T}}$, which consists of the preprocessing time required to compute the cost tables and the planning time required to generate the agents' paths. As shown in Figure~\ref{fig_r1k}, both the preprocessing time and the planning time grow approximately linearly with the number of agents. Even for the largest instances with 1{,}000 agents, PIBT$^{\text{*}}_{\text{T}}$ solves the problem within three seconds. This near-linear scaling suggests that the proposed approach is suitable for large-scale coordination scenarios and may be applicable to even larger instances. The average single-step planning time of PIBT$^{\text{*}}_{\text{T}}$ in the 1{,}000-agent case is 1.55\,ms, which is only 0.27\,ms higher than that of PIBT$_{\text{T}}$. This result indicates that the additional SPD and PET modules introduce only limited computational overhead.

\subsection{Gap between theoretical planning cost and realistic execution cost}
This subsection examines whether turn actions should be considered during high-level planning for realistic robot execution. As discussed in Section~\ref{intro}, solutions generated by classical MAPF algorithms ignore turning actions, but they can still be converted into conflict-free executable plans through low-level mechanisms such as ADG \citep{honig2019persistent}. This motivates a direct comparison between such a hierarchical approach and the proposed turn-aware high-level planning approach. Specifically, we compare the solutions generated by PIBT$^{\text{*}}_{\text{T}}$ and the classical MAPF solver PIBT. Both solutions are fed into an ADG-based simulator that incorporates turning actions, acceleration, and deceleration. The simulation results are used to quantify the gap between theoretical planning cost and actual execution cost, and to compare the actual execution performance of the two methods.

\begin{figure}[!htbp]
	\centering
	\includegraphics[width=0.85\linewidth]{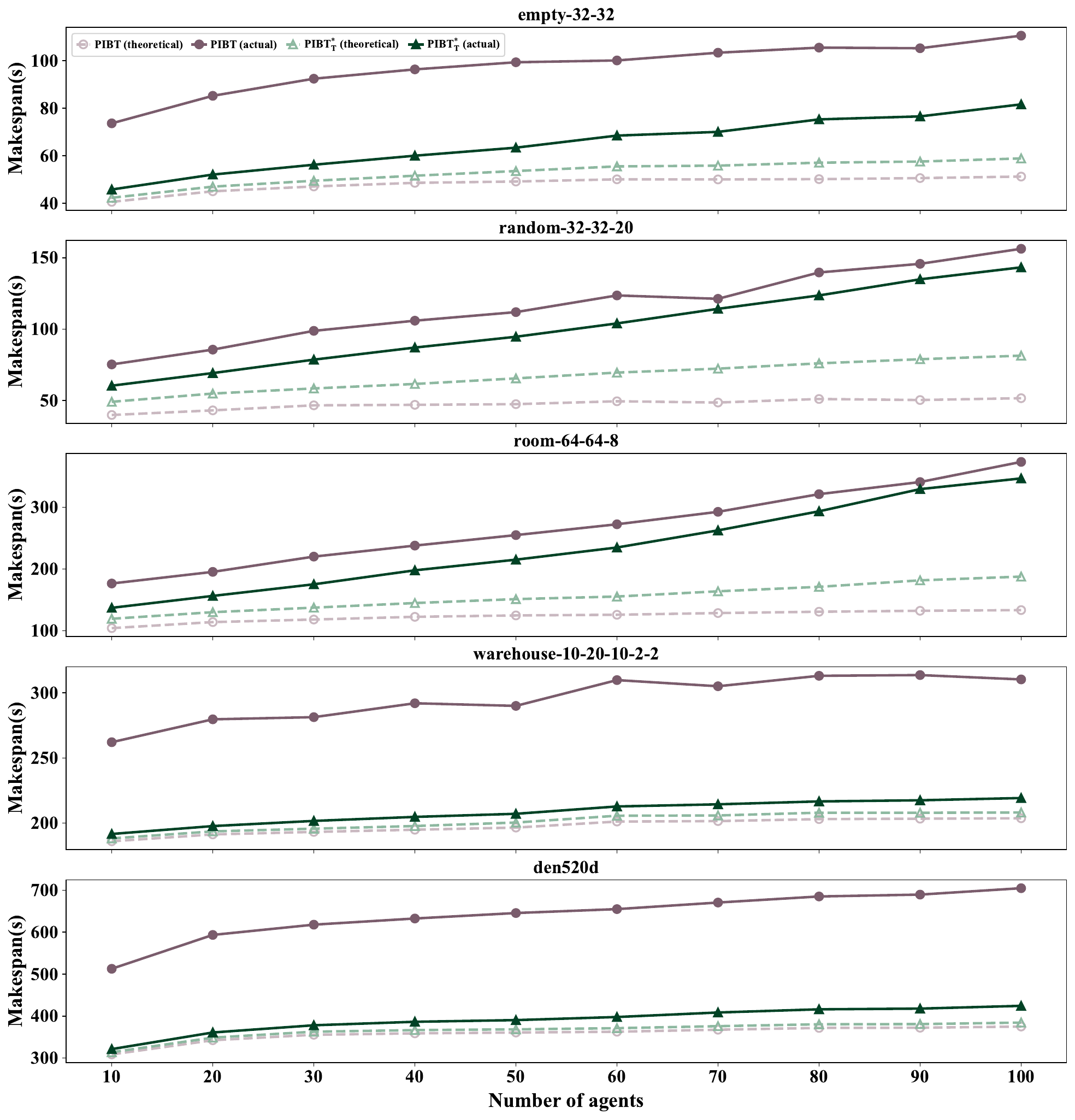}
	\caption{Comparison between theoretical planning makespan and ADG-based actual execution makespan for PIBT and PIBT$^{\text{*}}_{\text{T}}$.}
	\label{fig_ms_cp}
\end{figure}

Based on \citet{honig2019persistent}, we implement an ADG-based simulator that converts each discrete plan into a sequence of executable rotation, translation, and waiting actions. For each robot, Type-1 dependencies are generated to preserve the intra-robot action order along its planned path, while Type-2 dependencies are created according to the visiting order of shared grid cells among different robots. Since the converted ADG may contain cyclic dependency structures when multiple robots are mutually dependent within the same discrete planning step, we handle such cases through strict synchronization: mutually dependent actions are released simultaneously once all external predecessors have been completed. After the ADG is constructed, the simulator assigns continuous execution durations to translations and \(90^\circ\) rotations according to the robots' maximum velocity, acceleration, and angular velocity. In this way, the simulator captures the additional execution cost caused by turning, acceleration and deceleration, while preserving the dependency structure required for collision-free execution.

In the ADG-based simulation, each grid edge represents a physical distance of \(1\,\mathrm{m}\). All robots are assumed to have identical kinematic parameters. The maximum translational velocity is set to \(1\,\mathrm{m/s}\), the translational acceleration and deceleration are both set to \(1\,\mathrm{m/s^2}\), and the maximum angular velocity is set to \(1.5708\,\mathrm{rad/s}\). Consecutive forward actions in the same direction are merged into a single straight-line motion segment, and the robot does not stop at intermediate grid nodes. Under the specified parameters, a \(90^\circ\) in-place rotation requires approximately one second.

The numerical results are shown in Figure~\ref{fig_ms_cp}. PIBT$^{\text{*}}_{\text{T}}$ consistently achieves a smaller execution makespan than the paths directly generated by PIBT across all tested maps and agent numbers. This indicates that explicitly modeling turn actions during high-level planning improves the quality of the paths executed under realistic kinematic constraints. Moreover, the gap between theoretical planning cost and actual execution cost is much smaller for PIBT$^{\text{*}}_{\text{T}}$ than for PIBT. In contrast, although PIBT produces relatively low theoretical makespan values, its execution cost increases significantly after turn actions, acceleration/deceleration, and inter-robot dependencies are considered in the ADG-based simulator. These results demonstrate that classical MAPF solutions may substantially underestimate actual execution cost when realistic motion effects are ignored, whereas the proposed method produces solutions that are both more efficient in execution and more consistent with their theoretical cost estimates.

\subsection{Case study on a flexible manufacturing system}
\begin{figure}[tb]
	\centering
	\includegraphics[width=0.8\linewidth]{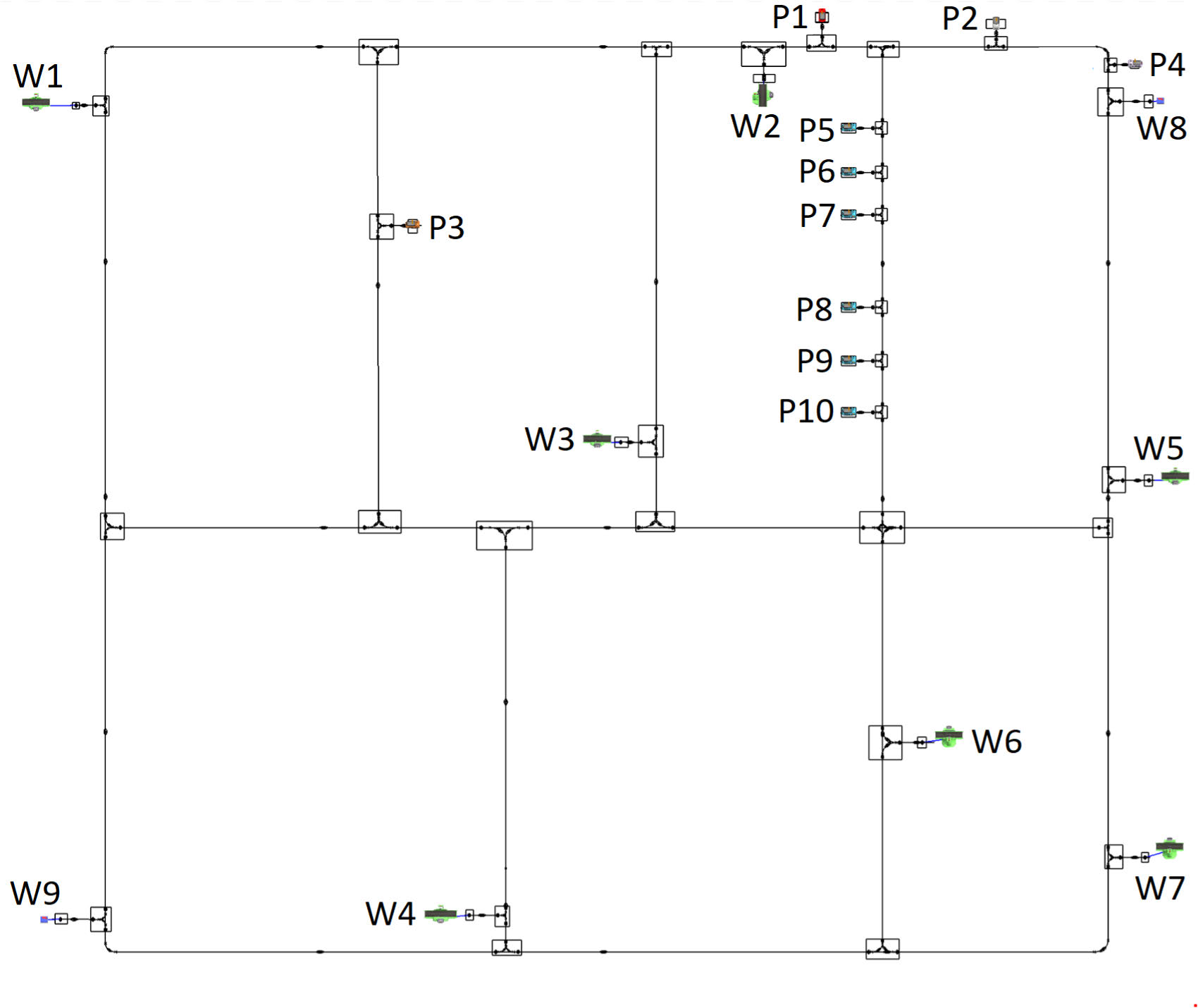}
	\caption{Example of a FMS layout, where P1 to P10 are parking stations and W1 to W9 are working stations.}
	\label{fig_fms}
\end{figure}

MMRSs are increasingly deployed in industrial environments. To evaluate the practical relevance of the proposed approach, this subsection presents a case study on pallet transportation in a Flexible Manufacturing System (FMS), adapted from \citet{verma2024traffic}. As illustrated in Figure~\ref{fig_fms}, the layout contains two types of functional areas: parking stations, labeled P, where idle robots may park and charge their batteries, and pick-up/drop-off stations, labeled W, where pallets are loaded or unloaded. Robots travel between these areas to perform logistics tasks. The map covers an area of \(44\times48\)\,m. The main corridors connecting the P and W stations are wide enough to accommodate two robots moving side by side, whereas the branch paths leading to individual P and W stations allow only one robot to pass at a time. This layout therefore contains both bidirectional traffic corridors and narrow single-lane passages, making it suitable for evaluating coordination performance under realistic industrial constraints.

To assess how the action sequences generated by different planners perform under realistic execution conditions, we employ the Scalable Multi-Agent Realistic Testbed (SMART), a state-of-the-art multi-robot simulation platform developed by \citet{yan2026advancing}. SMART provides a configurable execution environment for evaluating MAPF planners under robot kinematic constraints. 

The kinematic parameters of robots are configured as follows: the linear velocity of each robot is set to $1$\,m/s, the in-place rotation speed is set to $0.5\pi$\,rad/s, and the linear acceleration is set to $1$\,m/s\textsuperscript{2}. A total of ten instances are evaluated, of which five involve 10~agents and the remaining five involve 20~agents. To better simulate the variety of transport tasks encountered in a real FMS, the 
start and goal locations in each instance are drawn from a mixture of scenarios, including P-to-W, W-to-P, random-to-W, and random-to-P assignments. The baseline algorithm used for comparison is the original PIBT, which does not account for turn actions. By comparing the actual execution makespan of both algorithms under the SMART simulator, we analyze the gap between planners operating under different kinematic models and the resulting real-world execution efficiency.

\begin{figure}[tb]
	\centering
	\includegraphics[width=0.9\linewidth]{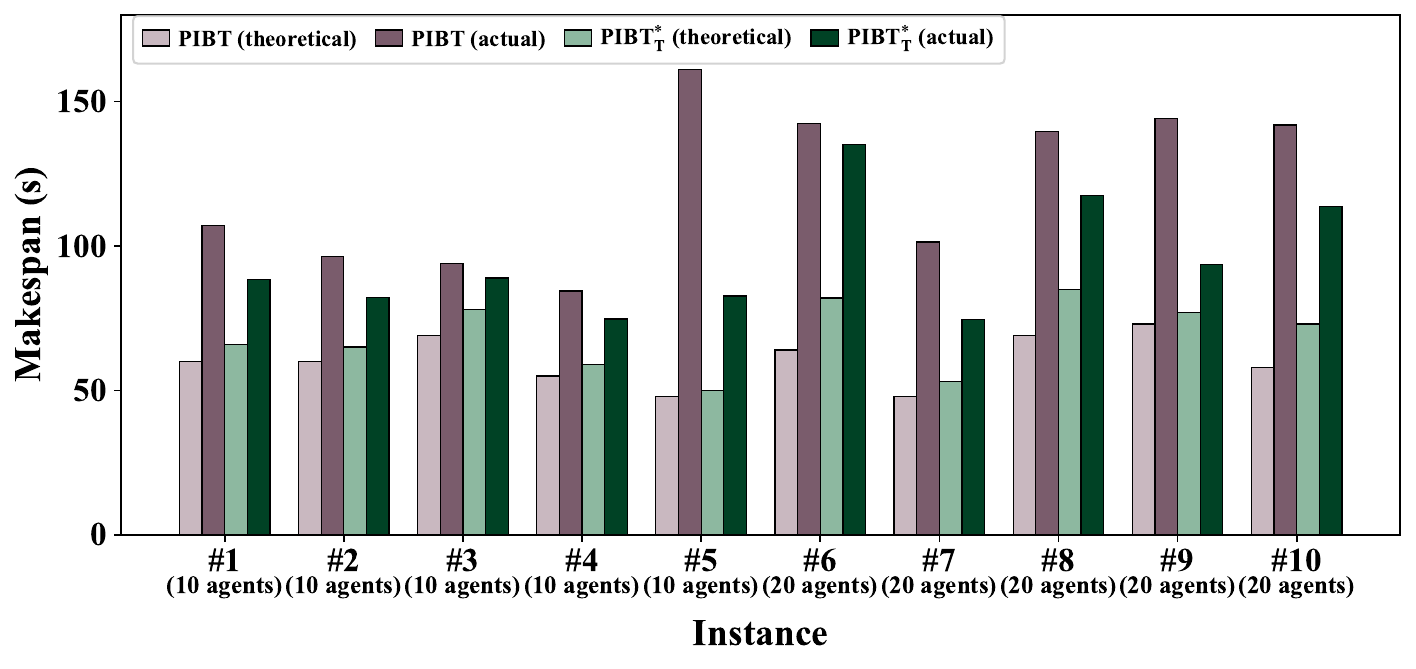}
	\caption{Comparison between theoretical planning makespan and actual execution makespan in SMART.}
	\label{fig_fms_a}
\end{figure}

% \begin{figure}[t]
% 	\centering
% 	\includegraphics[width=0.8\linewidth]{fig/fig_fms_path_small.png}
% 	\caption{Paths of three specific agents in SMART generated by PIBT (above) and PIBT$^{\text{*}}_{\text{T}}$ (below) of instance \#1 (10 agents).}
% 	\label{fig_fms_path}
% \end{figure}

\begin{figure}[tb]
	\centering
	\includegraphics[width=0.9\linewidth]{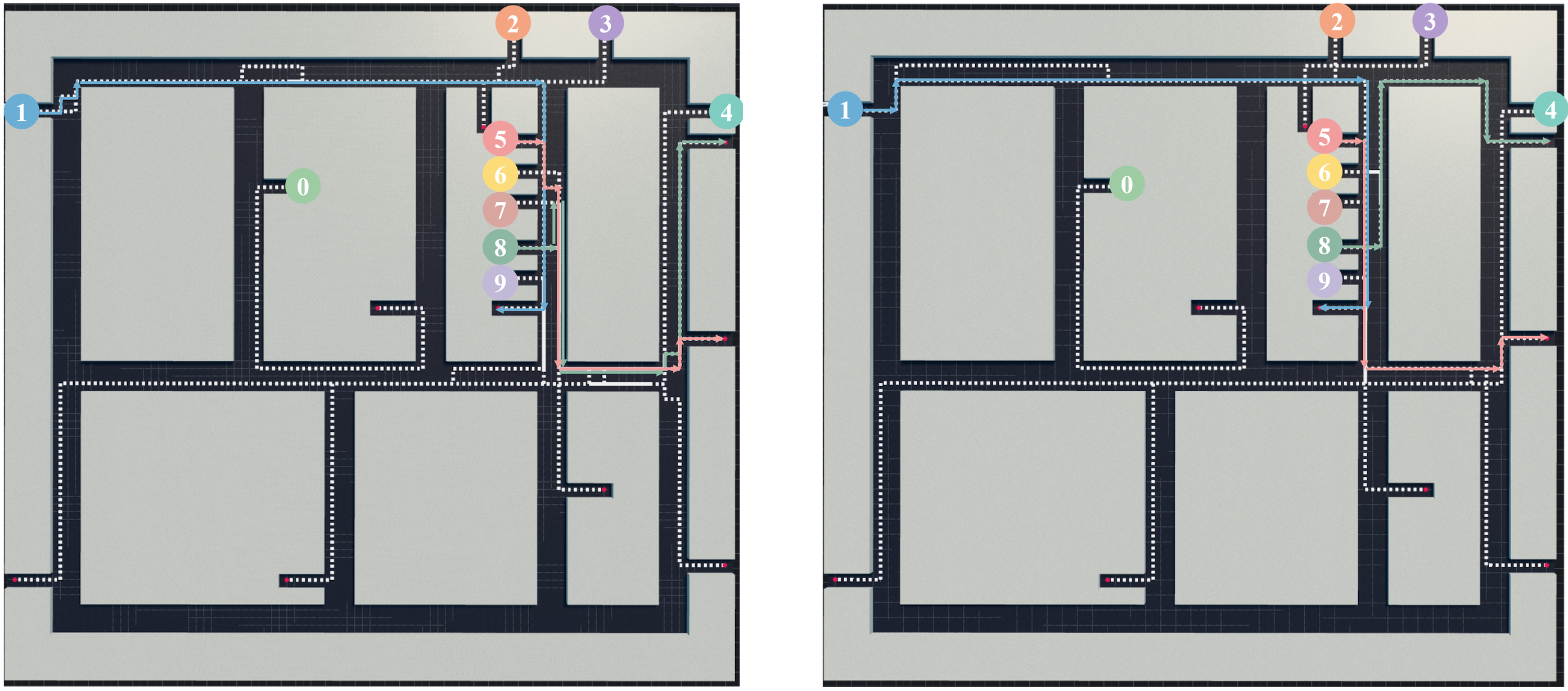}
	\caption{Execution paths of agents \(a_1\), \(a_5\), and \(a_8\) in the SMART simulator for instance \#1 with 10 agents. The left panel corresponds to PIBT, and the right panel corresponds to PIBT$^{\text{*}}_{\text{T}}$.}
	\label{fig_fms_path_new}
\end{figure}

Figure~\ref{fig_fms_a} reports the theoretical planning makespan and SMART-based actual execution makespan of PIBT and PIBT$^{\text{*}}_{\text{T}}$ across all ten FMS instances. Although PIBT$^{\text{*}}_{\text{T}}$ yields a higher theoretical planning makespan than PIBT, it achieves a lower actual execution makespan in every instance under SMART. On average, the actual execution makespan is reduced by 23.3\% for the 10-agent instances and by 20.2\% for the 20-agent instances. These results confirm that the advantage of turn-aware high-level planning is preserved in an industrially relevant FMS layout.

Figure~\ref{fig_fms_path_new} further explains this improvement by visualizing the execution paths of three representative agents in instance~\#1. Compared with PIBT, PIBT$^{\text{*}}_{\text{T}}$ produces smoother paths with fewer directional changes, because its turn-aware cost model penalizes unnecessary rotations during planning. This directly reduces the execution delay caused by in-place turns. In addition, the figure reveals an inefficiency caused by repeated pushing in priority-based planning. Under PIBT, agent \(a_8\) initially moves upward after leaving the garage, but is then repeatedly pushed downward by the higher-priority agent \(a_5\), eventually reaching its goal through a substantially longer detour. PIBT$^{\text{*}}_{\text{T}}$ mitigates this behavior and guides the agents through more efficient execution paths. A simulation video is provided in the supplementary material.

The FMS case study therefore complements the benchmark-based ADG simulation in the previous subsection. While the benchmark experiments quantify the general gap between theoretical planning cost and realistic execution cost, the FMS case study shows how this gap translates into practical execution inefficiency in a structured industrial layout. The results indicate that PIBT$^{\text{*}}_{\text{T}}$ not only reduces unnecessary rotations, but also alleviates repeated pushing and detours, leading to more efficient robot traffic in realistic manufacturing scenarios.

\section{Demonstration with real robots} \label{sec:real_robot}
\begin{figure}[tb]
	\centering
	\includegraphics[width=\linewidth]{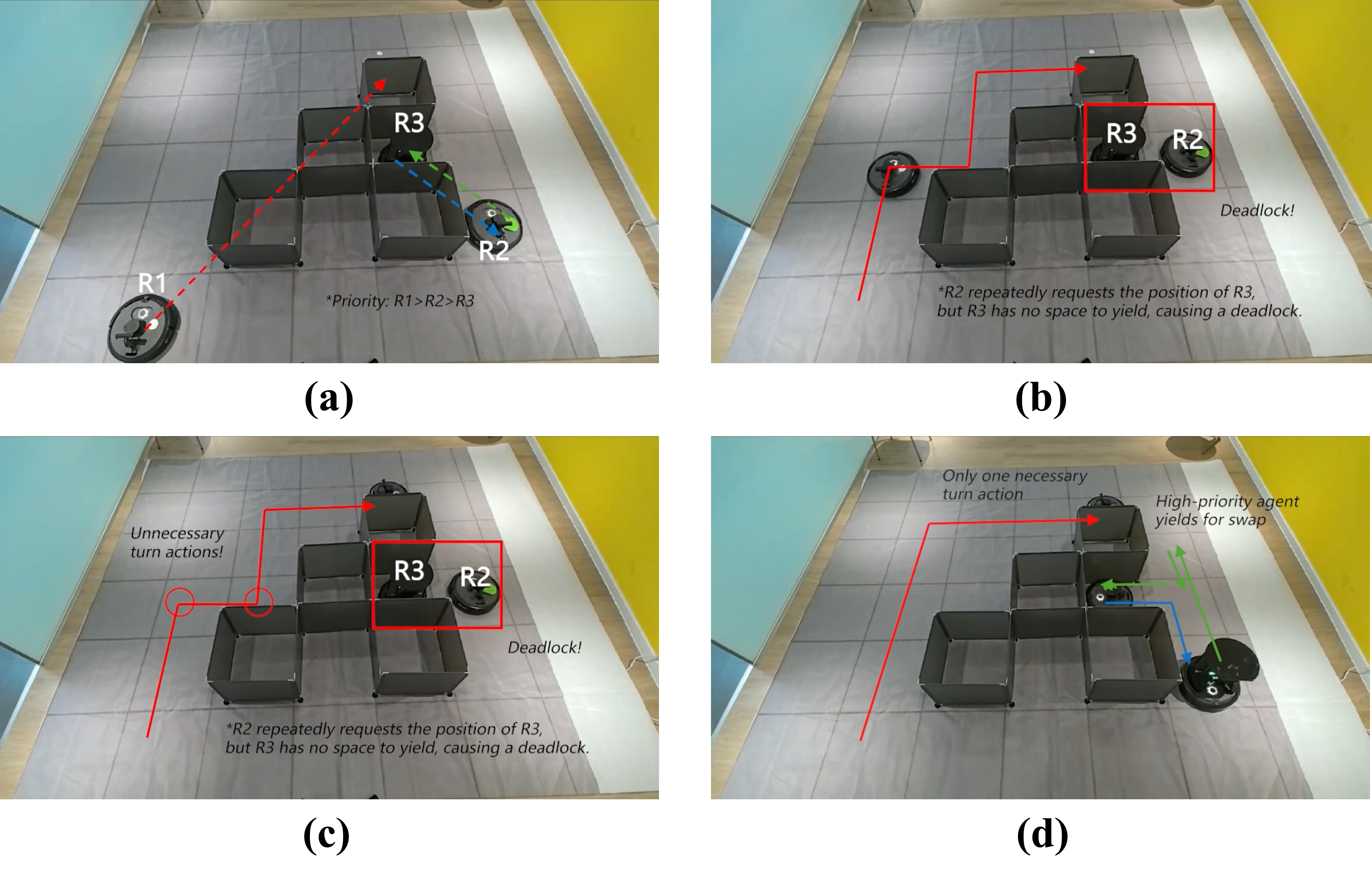}
	\caption{Real-robot experiment on three TurtleBot 4 robots.
        (a) Start and goal configuration on the $6\times6$ map with obstacles.
        (b) Deadlock between R2 and R3 under PIBT+ADG at $t=5$\,s.
        (c) R1 reaches its goal under PIBT+ADG at $t=41$\,s, while R2 and R3 remain deadlocked.
        (d) Successful execution under PIBT$^{\text{*}}_{\text{T}}$: R1 reaches its goal at $t=32$\,s, and R2 and R3 reach their goals at approximately $t=36$\,s.}
	\label{fig_tb4}
\end{figure}

To further validate the applicability of the proposed algorithm on real robots, we conduct a physical experiment using three TurtleBot4 robots on a $6\times6$ grid map with obstacles. The robot team consists of two TurtleBot4 Lite robots and one TurtleBot4 Standard robot, denoted as R1, R2, and R3, respectively. The start and goal locations of the three robots are shown in Figure~\ref{fig_tb4}(a). In the experiment, the high-level planner first computes a discrete path plan for each robot. The resulting plans are then processed by an ADG-based controller, which constructs the action dependency relationships among robots and converts the discrete plans into adjusted continuous action sequences. These action sequences are subsequently distributed to the corresponding robots and executed on the physical platform.

We compare two execution pipelines on the same task: PIBT combined with ADG-based execution, and the proposed PIBT$^{\text{*}}_{\text{T}}$ combined with the same ADG-based execution framework. In the first pipeline, the paths generated by PIBT are directly passed to the ADG-based controller. Although the controller preserves collision-free execution, the resulting execution can still suffer from poor coordination quality. As shown in Figure~\ref{fig_tb4}(b), R2 and R3 enter a deadlock approximately 5 seconds after the start of execution and remain stationary thereafter. This deadlock occurs because R2 repeatedly requests the position occupied by R3, while R3 has no available space to yield. Consequently, the two robots mutually block each other and fail to complete their tasks. In the same experiment, R1 eventually reaches its goal after 41 seconds, as shown in Figure~\ref{fig_tb4}(c). However, its action sequence contains three turning actions, two of which are unnecessary, thereby increasing the execution time.

In contrast, when PIBT$^{\text{*}}_{\text{T}}$ is used as the high-level planner, all three robots successfully reach their goal locations. As shown in Figure~\ref{fig_tb4}(d), R1 reaches its goal in 32 seconds, which is 9 seconds faster than in the PIBT+ADG setting. This improvement is mainly because PIBT$^{\text{*}}_{\text{T}}$ explicitly accounts for turn actions during planning and avoids unnecessary orientation changes. Meanwhile, R2 successfully yields to R3, allowing both robots to pass through the constrained area and reach their respective goals approximately 36 seconds after the start of execution. These results show that the proposed method not only improves execution efficiency, but also avoids deadlock patterns that may arise when turn actions and their induced delays are ignored at the planning level. The complete experimental video is provided in the supplementary material.

\section{Conclusion}
\label{sec:conclusion}

This paper presents PIBT$^{\text{*}}_{\text{T}}$, an efficient priority-inheritance-based path planning method for large-scale multi-mobile-robot systems under turn constraints. By extending the original PIBT framework to an orientation-aware setting, the proposed method preserves the computational efficiency of single-timestep priority-inheritance planning while addressing the key difficulties introduced by turn actions. To this end, we develop several dedicated mechanisms, including turn-aware action updates, cycle handling for circular request chains, a swap pattern detector (SPD) for resolving local deadlocks, and a \(k\)-Push Escape Trigger (PET) for mitigating livelock and inefficient repeated pushing behaviors.

Extensive computational experiments are conducted to evaluate the proposed method. First, the impact of the PET threshold parameter \(k\) and the stability of the algorithm outputs are systematically analyzed. Second, the effects of the proposed SPD and PET modules are investigated through an ablation study. Third, the proposed method is compared with representative MAPF\textsubscript{T} solvers on standard benchmarks. The results show that PIBT$^{\text{*}}_{\text{T}}$ achieves high success rates and strong scalability. Furthermore, a large-scale stress test shows that the proposed method can solve instances with up to 1{,}000 agents within seconds while incurring only limited computational overhead.

To investigate the execution performance of the proposed method, we conduct an ADG-based execution simulation, in which the solutions generated by different planners are converted into executable action sequences subject to kinematic constraints. Compared with classical MAPF approaches that ignore turn actions, PIBT$^{\text{*}}_{\text{T}}$ produces plans whose estimated costs are more consistent with realistic execution and achieves lower actual execution makespan. We further validate the practical performance of the proposed method through a case study on a flexible manufacturing system, where AGVs are used to perform logistics tasks. Finally, a real-robot demonstration with TurtleBot4 robots further confirms the feasibility of deploying the proposed method on physical multi-robot platforms.

Future work will first focus on strengthening the theoretical foundations of the proposed method. Although PIBT$^{\text{*}}_{\text{T}}$ achieves strong empirical performance and scalability, it remains a heuristic method and does not yet provide formal completeness or optimality guarantees. Establishing stronger guarantees on reachability, termination, and solution quality under turn constraints is therefore an important direction for future research. Another promising direction is to incorporate learning-based mechanisms into priority-inheritance-based planning, for example to improve node selection, escape decisions, and conflict-resolution strategies while preserving the scalability and interpretability of the proposed framework. Finally, we plan to extend the proposed method to more realistic multi-robot coordination systems with dynamic tasks, uncertain execution times, and online disturbances, such as integrated task assignment and path planning, execution-aware replanning under real-time feedback, and lifelong coordination of robot fleets in logistics and manufacturing environments.

\section*{Acknowledgements}
The research was supported by the National Natural Science Foundation of China [Grant No.72301196] and the Fundamental Research Funds for the Central Universities, Tongji University [Grant No.22120240607].

\bibliographystyle{elsarticle-harv}
\bibliography{reference}

\end{document}